\documentclass[sigconf]{acmart}
\AtBeginDocument{%
  }

\setcopyright{acmlicensed}
\copyrightyear{2026}
\acmYear{2026}
\setcopyright{cc}
\setcctype{by}
\acmConference[KDD '26]{Proceedings of the 32nd ACM SIGKDD Conference on Knowledge Discovery and Data Mining V.2}{August 09--13, 2026}{Jeju Island, Republic of Korea}
\acmBooktitle{Proceedings of the 32nd ACM SIGKDD Conference on Knowledge Discovery and Data Mining V.2 (KDD '26), August 09--13, 2026, Jeju Island, Republic of Korea}
\acmDOI{10.1145/3770855.3817926}
\acmISBN{979-8-4007-2259-2/2026/08}
\usepackage[table]{xcolor}
\usepackage{multirow}
\usepackage{wrapfig}
\usepackage{tcolorbox}
\usepackage{balance}

\definecolor{myred}{RGB}{0,0,0}
\begin{document}

\title{Parametric Social Identity Injection and Diversification in Public Opinion Simulation}


\author{Hexi Wang}
\affiliation{%
  \institution{DCST, Tsinghua University}
  \city{Beijing}
  \country{China}}
\affiliation{%
  \institution{Quancheng Laboratory}
  \city{Jinan}
  \state{Shandong}
  \country{China}}
\email{whx25@mails.tsinghua.edu.cn}
\orcid{0009-0002-0763-2433}


\author{Yujia Zhou}
\authornote{Corresponding authors.}
\affiliation{%
  \institution{Quancheng Laboratory}
  \city{Jinan}
  \state{Shandong}
  \country{China}}
\affiliation{%
  \institution{DCST, Tsinghua University}
  \city{Beijing}
  \country{China}}
\email{zhouyujia@mail.tsinghua.edu.cn}
\orcid{0000-0002-3530-3787}

\author{Bangde Du}
\affiliation{%
  \institution{DCST, Tsinghua University}
  \city{Beijing}
  \country{China}}
\email{dbd23@mails.tsinghua.edu.cn}
\orcid{0009-0001-0212-5136}


\author{Qingyao Ai}
\authornotemark[1]
\affiliation{%
  \institution{Quancheng Laboratory}
  \city{Jinan}
  \state{Shandong}
  \country{China}}
\affiliation{%
  \institution{DCST, Tsinghua University}
  \city{Beijing}
  \country{China}}
\email{aiqy@tsinghua.edu.cn}
\orcid{0000-0002-5030-709X}

\author{Yiqun Liu}
\affiliation{%
  \institution{DCST, Tsinghua University}
  \city{Beijing}
  \country{China}}
\email{yiqunliu@tsinghua.edu.cn}
\orcid{0000-0002-0140-4512}


\begin{abstract}
  Large language models (LLMs) have recently been adopted as synthetic agents for public opinion simulation, offering a promising alternative to costly and slow human surveys. Despite their scalability, current LLM-based simulation methods fail to capture social diversity, producing flattened inter-group differences and overly homogeneous responses \textcolor{myred}{across} demographic groups. We identify this limitation as a Diversity Collapse phenomenon in LLM hidden representations, where distinct social identities become increasingly indistinguishable across layers. Motivated by this observation, we propose Parametric Social Identity Injection (PSII), a general framework that injects explicit, parametric representations of demographic attributes and value orientations directly into intermediate hidden states of LLMs. Unlike prompt-based persona conditioning, PSII enables fine-grained and controllable identity modulation at the representation level. Extensive experiments on the World Values Survey using multiple open-source LLMs show that PSII significantly improves distributional fidelity and diversity, reducing KL divergence to real-world survey data while enhancing overall diversity. This work provides new insights into representation-level control of LLM agents and advances scalable, diversity-aware public opinion simulation.
\end{abstract}




\maketitle
\newcommand\kddavailabilityurl{https://doi.org/10.5281/zenodo.20465632}
\ifdefempty{\kddavailabilityurl}{}{
\begingroup\small\noindent\raggedright\textbf{Resource Availability:}\\
The source code and data used in this paper are publicly available at \url{https://github.com/halsayxi/PSII}, with a versioned archival release available at \url{\kddavailabilityurl}.
\endgroup
}


\section{Introduction}

Public opinion simulation~\cite{groves2011survey} is critical for quantifying societal attitudes, yet traditional surveys face escalating costs and scalability issues~\cite{dillman2014internet, groves2010total, tourangeau2000psychology, krumpal2013determinants}. To address this, recent work explores \textbf{agent-based modeling} (ABM) using \textbf{large language models} (LLMs) as synthetic respondents. Leveraging LLMs enables efficient, low-cost simulations with several advantages, including but not limited to reduced logistical burdens, flexible experimental scenario design, unlimited follow-ups, and multi-dimensional and controllable experimental populations conditioned on demographic, socioeconomic, or ideological attributes.

\begin{figure*}[ht]
  \centering
  \includegraphics[width=\linewidth]{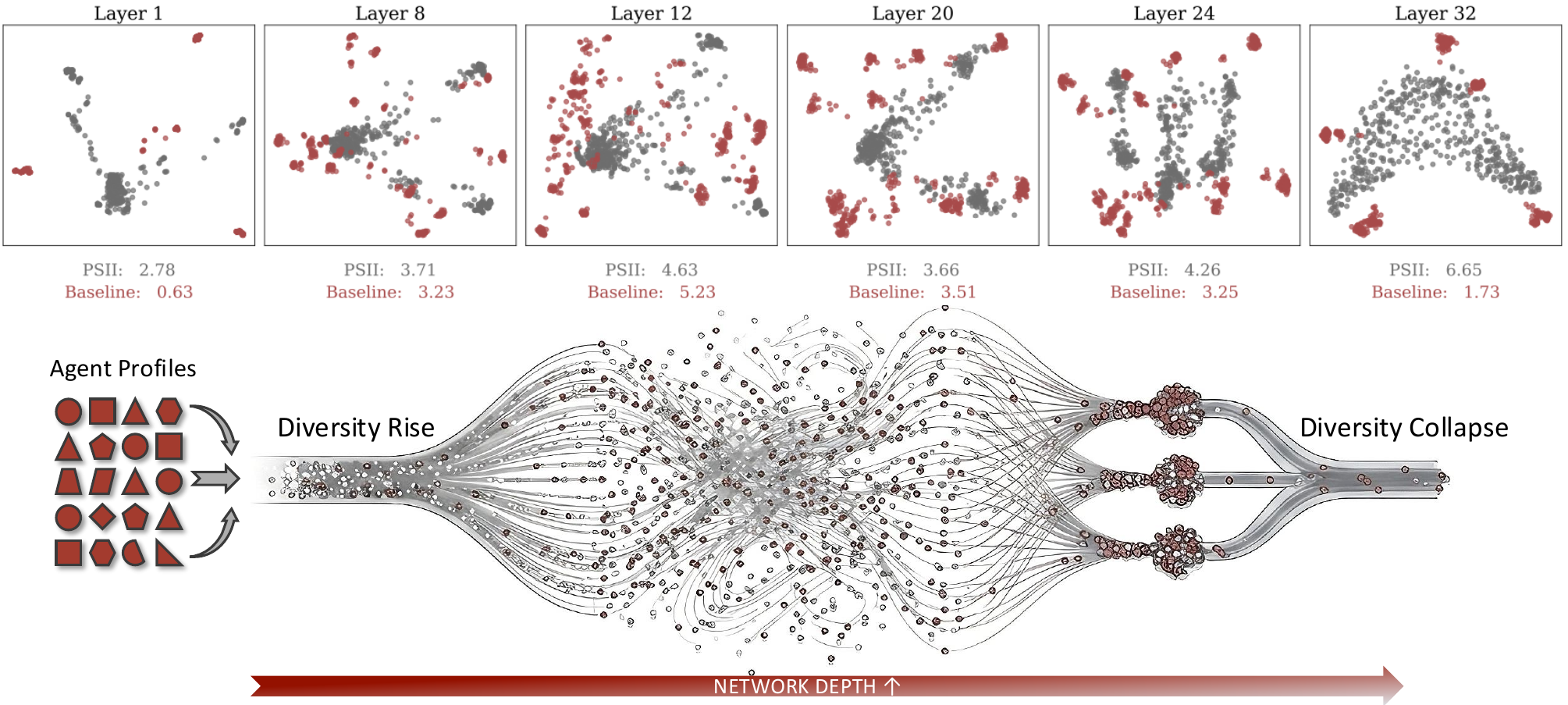}
  \caption{Layer-wise scatter plots of final-token hidden states for 500 simulated agents (top) and an illustration of Diversity Collapse in Transformer hidden states (bottom).
In the top panels, red points denote baseline methods and gray points denote PSII-generated agents; the reported scores measure the average spatial dispersion of representations in each layer.
The bottom panel depicts the Diversity Collapse phenomenon.}
  \label{fig:merge}
  \Description{merge}
\end{figure*}

While LLM-based opinion simulation approaches demonstrate promising performance on specific tasks, they often share a critical limitation: insufficient diversity in simulated populations. 
Diversity is crucial for social research and public opinion studies. \textcolor{myred}{Even small degrees of population homogenization or representational bias can lead to misleading conclusions about societal dynamics}~\cite{myers2021rooting, hemmatian2022debiased, qu2024performance, karanjai2025synthesizing, fabris2022algorithmic, shumailov2024ai}. 
Yet, as shown in our experiments and previous studies~\cite{bisbee2024synthetic, kaiser2025simulating, kitadai2024examining, park2024diminished, boelaert2025machine}, \textcolor{myred}{existing LLM-based approaches often produce homogeneous results whose behavioral distributions differ significantly from those of real human subjects.} LLM-based approaches to public opinion simulation from previous studies generally can be categorized as direct zero-shot querying~\cite{santurkar2023whose}, persona-based prompting~\cite{yang-etal-2024-large-language-models-llms, hwang-etal-2023-aligning, beck-etal-2024-sensitivity, yang2024large, du-etal-2025-simvbg}, or fine-tuning based alignment~\cite{suh-etal-2025-language, wang2024large, huang2025distribution}. 
\textcolor{myred}{Their limitations in diversity manifest at two distinct levels.}
First, the \textbf{inter-group diversity}. 
Standard training objectives for LLMs, such as maximum likelihood estimation with cross-entropy loss, inherently favor the most probable continuations. As a result, minority or low-frequency viewpoints tend to be underrepresented, leading to flattened response distributions in which distinct subpopulations become difficult to distinguish~\cite{wang2025large}.
Second, \textbf{intra-group diversity}. When demographic attributes are injected via prompts, identities are treated as fixed explanatory variables that dominate response generation~\cite{wang2025large}. This method ignores the heterogeneity within groups, exaggerates between-group differences, and reinforces group stereotypes as a consequence. 

To understand diversity loss in LLM agents, we analyze \textbf{internal hidden-state representations}. We projected final-token hidden states from 500 agents using KPCA to visualize representational diversity across layers (Figure~\ref{fig:merge}). From the red baseline points, our analysis reveals a non-monotonic pattern: lower layers form compact clusters, while intermediate layers spread out, reaching high diversity. Critically, higher layers experience a systematic contraction, collapsing into dense clusters, a phenomenon we term \textbf{Diversity Collapse}. The above analysis highlights a fundamental limitation of existing approaches: they lack stable and heterogeneous conditions to guide hidden-state evolution throughout the network. \textcolor{myred}{See Appendix~\ref{sec:addres3} for details.}

The analysis highlights that existing approaches lack stable conditions to guide hidden-state evolution. As external inputs, textual prompts are progressively smoothed across layers, making them insufficient for sustaining structured individual differences. Consequently, synthetic agents often possess weakly grounded identity representations. Motivated by these observations, we propose \textbf{Parametric Social Identity Injection (PSII)}, which explicitly models social identity within the internal representation space. 
Similar to Parametric RAG~\cite{10.1145/3726302.3729957}, PSII embeds identity information as parametric vectors, including \textbf{demographic and value vectors}, directly into hidden states.
This enables identity attributes to shape hidden-state trajectories rather than relying on surface prompts. To enhance \textbf{inter-group diversity}, PSII introduces stable signals that persist across layers; to preserve \textbf{intra-group diversity}, we apply stochastic perturbations to simulate natural variation. Beyond effectiveness, PSII offers three key advantages: \textbf{efficiency}, using vectors to modulate identities with minimal storage and no fine-tuning; \textbf{reusability}, which forms a modular library of agent identities applicable across various datasets and enables the efficient generation of diverse synthetic populations at a scale proportional to the Cartesian product of attribute dimensions; and \textbf{tractability}, enabling precise quantitative analysis via linear algebraic operations.

We evaluate PSII on the World Values Survey (WVS)\footnote{\url{https://www.worldvaluessurvey.org/wvs.jsp}} dataset using multiple open-source LLMs, including Qwen2.5-7B-Instruct and Qwen2.5-14B-Instruct~\cite{bai2023qwen}, Llama-3.1-8B-Instruct~\cite{touvron2023llama}, and Mistral-24B-Instruct~\cite{mistral_small3_2025}. Across models and tasks, PSII consistently improves both prediction accuracy with respect to human responses and diversity metrics. Notably, in the same layer-wise visualization (Figure~\ref{fig:merge}, top), the gray points show representations produced by PSII, which exhibit sustained or even increasing diversity in higher layers, effectively counteracting the Diversity Collapse observed in baseline methods.

In summary, this work makes three primary contributions. First, we identify and characterize the Diversity Collapse phenomenon in the hidden states of LLM-based social simulation agents, which explains why conventional prompt-based methods fail to capture population heterogeneity. Second, we propose Parametric Social Identity Injection (PSII), a principled framework for stable and heterogeneous identity modeling by injecting identity vectors into hidden representations. Third, through systematic experiments on WVS, we demonstrate that PSII significantly improves diversity and distributional fidelity, producing synthetic populations that better reflect real-world heterogeneity.

\section{\textcolor{myred}{Related Work}}

\subsection{LLM-based Personality Simulation Agents}

Early research focused on LLMs' zero-shot capabilities. \textcolor{myred}{While direct prompting of models on specific topics is a fundamental baseline}, studies show default outputs often exhibit political biases (e.g., left-leaning) and underrepresent marginalized groups~\cite{argyle2023out, santurkar2023whose}.

To improve realism, Persona-based Prompting injects demographic attributes (age, gender, race) into prompts~\cite{yang-etal-2024-large-language-models-llms, zhou2505investigating, yang2024large}. Hwang et al.~\cite{hwang-etal-2023-aligning} proposed a framework for aligning with user opinions, demonstrating that persona-based personalized prompting significantly improves prediction accuracy. However, Beck et al.~\cite{beck-etal-2024-sensitivity} pointed out in their study that Sociodemographic Prompting involves a trade-off between sensitivity and robustness, and simple attribute injection may lead to model stereotyping.

Recent trends shift toward Fine-tuning and Alignment~\cite{suh-etal-2025-language, wang2024large}. Compared to general-purpose models, models fine-tuned on specific social survey data exhibit stronger distribution fitting capabilities. For instance, SimVBG simulates complex values via individual backstories~\cite{du-etal-2025-simvbg}, while Distribution Shift Alignment (DSA) helps models adapt to context changes~\cite{huang2025distribution}.

To better mirror real-world demographics, Chen et al.~\cite{chen2026hag} proposed HAG (Hierarchical Demographic Tree-based Agent Generation), utilizing a hierarchical tree structure to generate topic-adaptive agents. Hu et al.~\cite{hu2025population} introduced Population-Aligned Persona Generation, utilizing importance sampling to reduce population-level biases. Chen et al.~\cite{chen2025persona} further explored precisely regulating model personality traits by controlling specific directions in the activation space, providing a new technical pathway for high-fidelity social simulation.

\subsection{Enhancing Diversity and Representativeness}

Lack of diversity remains a challenge, where models favor generic opinions over minority voices.

\textcolor{myred}{Traditional sampling strategies, such as high-temperature or top-k sampling}, can increase randomness but often come at the cost of response coherence~\cite{platt1999probabilistic, chung2023increasing}. To enhance diversity while maintaining quality, Wong et al.~\cite{wong2024simplestrat} proposed SimpleStrat, leveraging the concept of stratification to guide the model in exploring different solution spaces. Zhang et al.~\cite{zhang2025cultivating} introduced Negatively-Correlated (NC) Sampling, which forces the model to unearth differentiated perspectives by suppressing the probability of already generated opinions, and released the Community Alignment Dataset to support research on pluralistic preferences.

In Prompt Engineering, mechanisms like Step-by-step Recall and Collective-Critique (CCSV) enhance viewpoint coverage and cultural diversity~\cite{hayati2024far, lahoti2023improving}. Multilingual Prompting also serves as an implicit cue to activate embedded cultural knowledge~\cite{wang2025multilingual}. Finally, research by Abels et al.~\cite{abels2025wisdom} suggests that relying solely on LLM populations may exacerbate biases. They proposed the concept of Hybrid Human-LLM Crowds, demonstrating that combining human diversity with LLM reasoning capabilities is an effective approach to mitigate bias and enhance collective intelligence.

\section{Parametric Social Identity Injection}

\begin{figure*}[ht]
  \centering
  \includegraphics[width=0.9\linewidth]{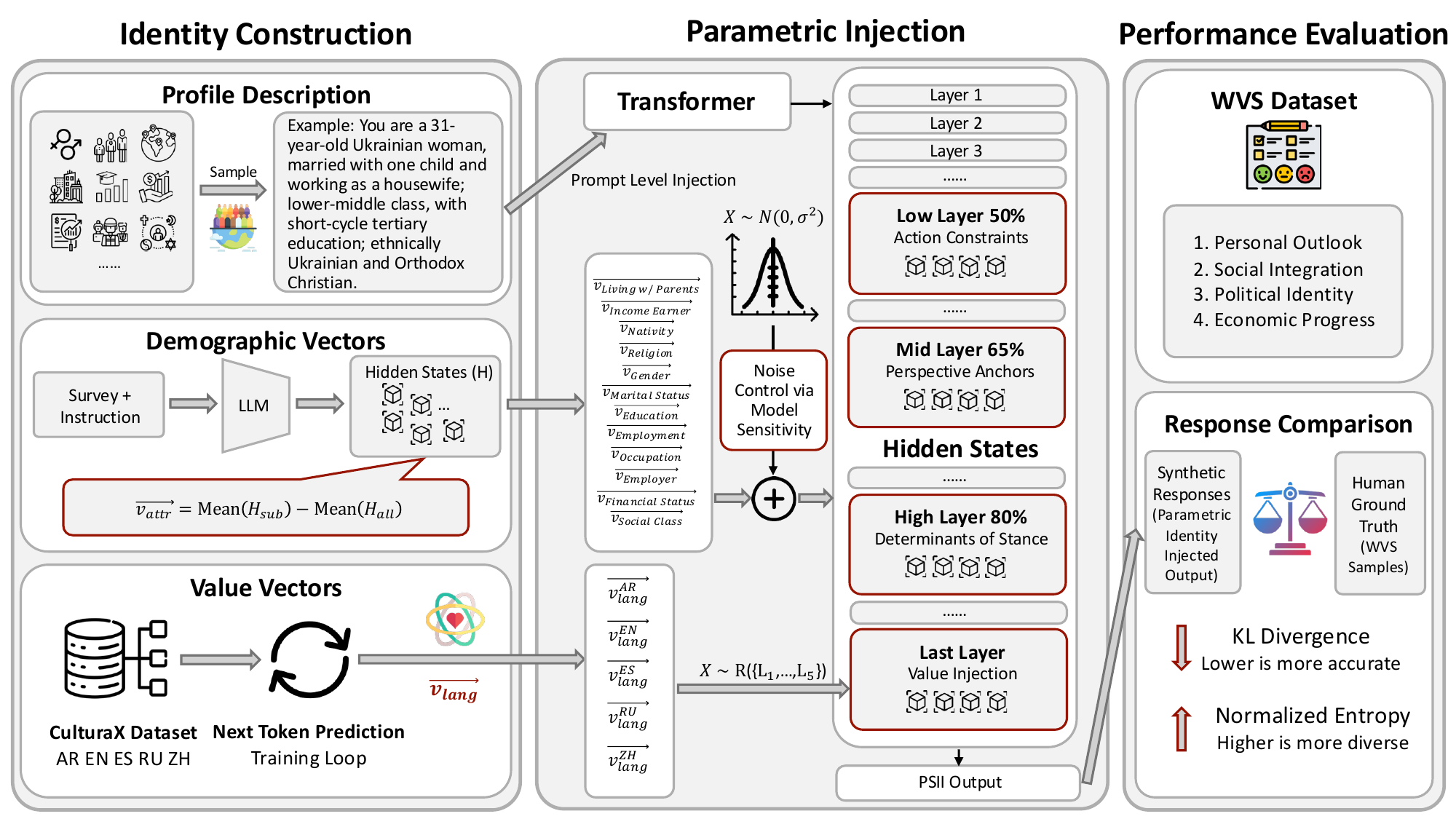}
  \caption{Overview of the Parametric Social Identity Injection (PSII) mechanism. From left to right: 
  \textbf{Identity Construction}, including agent profile construction and identity vector construction; 
  \textbf{Parametric Injection}, including noise addition and hierarchical injection; 
  \textbf{Performance Evaluation} of the simulated agents.}
  \label{fig:framework}
  \Description{Illustration of the PSII framework showing the pipeline from Identity Construction (agent profile and identity vector creation), through Parametric Injection (adding noise and hierarchical injection into hidden states), to Performance Evaluation of the simulated agents.}
\end{figure*}

We propose \textbf{Parametric Social Identity Injection (PSII)}, a framework for enhancing both fidelity and diversity in LLM-based public opinion simulation. As illustrated in Figure~\ref{fig:framework}, PSII injects structured identity information, including demographic and value-related features, directly into the model’s hidden states, introducing stable and heterogeneous individual differences that guide the generation of synthetic responses.

\subsection{Identity Construction}

PSII integrates two complementary components to model individual agents: \textbf{agent profiles}~(prompt level) and \textbf{identity vectors}~(hidden state level). This dual-level approach addresses the lack of stable personality modeling in conventional LLM simulations.

\subsubsection{Agent Profile Description}

For each synthetic agent, we construct a semantic agent profile using demographic variables. These variables are converted into descriptive text prompts, providing the model with semantic priors about the agent’s demographic context. Formally, let $P_i$ denote the agent profile of agent $i$, composed of a set of descriptive phrases:

\[
P_i = \{ d_1, d_2, \dots, d_M \},
\]

where $d_j$ is the textual description corresponding to the $j$-th demographic variable, and $M$ is the number of variables included. These profiles are used to condition the LLM at the \textbf{prompt level}, guiding its initial understanding of the agent’s characteristics. The specific prompts and the demographic features used are detailed in Appendix~\ref{app:baseline}.

\subsubsection{Demographic Vectors}

\textcolor{myred}{To enable stable and structured identity representation, we select representative demographic features that are broadly available across major social surveys, have stable definitions, and capture fundamental social positions, and then construct \textcolor{myred}{a} demographic vector for each feature value.} The specific features used are detailed in Appendix~\ref{app:dataset3}. Let $\mathcal{V}_k$ denote the set of possible values for demographic variable $k$, and $v_{k,j} \in \mathcal{V}_k$ a specific value. The construction process is as follows:

\textbf{Survey Question Simulation:} For each demographic variable $k$, we first define a fixed set of survey questions
$\{Q_k^{(1)}, \dots, Q_k^{(R)}\}$ that probe the semantic implications of this attribute.
For each value code $v_{k,j} \in V_k$, we then construct a set of value-specific prompts to elicit the model's internal representation of this identity.
By combining the shared question set with these value-specific instructions, we generate
a collection of synthetic prompts:
\[
\{(Q_k^{(r)}, v_{k,j}^{(m)}) \mid r = 1, \dots, R;\; m = 1, \dots, M_{k,j}\},
\]
where $v_{k,j}^{(m)}$ denotes the $m$-th role instruction associated with value
$v_{k,j}$, and $M_{k,j}$ is the number of role instructions defined for that value.
This results in $R \times \sum_j M_{k,j}$ question--response instances for each
demographic variable $k$, forming the demographic vector dataset, which is detailed in Appendix~\ref{app:dv_dataset}.

\textbf{LLM Response Embedding:}
For each synthetic prompt generated for demographic variable $k$ and value
$v_{k,j}$, the LLM produces a response.
Let $h^{(l)}_{k,j,m,t}$ denote the hidden state of token $t$ at layer $l$
for the response generated from the $m$-th role instruction associated with
value $v_{k,j}$.
We compute the \textbf{layer-wise average hidden state} for each response as:
\[
\bar{h}^{(l)}_{k,j,m} = \frac{1}{T_{k,j,m}} \sum_{t=1}^{T_{k,j,m}} h^{(l)}_{k,j,m,t},
\]
where $T_{k,j,m}$ is the number of tokens in the corresponding response.

\textbf{Value-Specific Vector Computation:}
The demographic vector $d_{k,j}$ is calculated as the difference
between the mean response representation over all prompts conditioned on $v_{k,j}$
and the marginal mean over all values of the same demographic variable $k$:
\[
\mathbf{d}_{k,j}
=
\frac{1}{|\mathcal{S}_{k,j}|}
\sum_{(k,j,m) \in \mathcal{S}_{k,j}}
\bar{h}^{(\mathcal{L}_k)}_{k,j,m}
-
\frac{1}{|\mathcal{S}_{k}|}
\sum_{(k,j',m') \in \mathcal{S}_{k}}
\bar{h}^{(\mathcal{L}_k)}_{k,j',m'},
\]
where $\mathcal{S}_{k,j}$ denotes the set of all response instances generated from
prompts conditioned on value $v_{k,j}$, $\mathcal{S}_{k} = \bigcup_{j} \mathcal{S}_{k,j}$ represents the union of these sets across all values of demographic variable $k$, and $\mathcal{L}_k$ is the layer selected for identity injection.

\textcolor{myred}{Additional analyses show that demographic-vector construction is robust across instruction models, prompt variants, and random seeds, and that the generated demographic semantics remain highly consistent across settings; see Appendix~\ref{app:vector_robustness}.}

\subsubsection{Value Vectors}

Language not only encodes information but also reflects cultural and worldview-specific patterns.

To approximate \textbf{value orientations} in the absence of explicit value annotations, we construct \textbf{language-based value vectors} using data from the target populations. \textcolor{myred}{Specifically, we learn a lightweight trainable vector $\mathbf{l}_s$ for each representative language $s$. Each value vector has the same dimensionality as the model hidden states, while all parameters of the base LLM are frozen. During training, $\mathbf{l}_s$ is optimized with the standard next-token prediction objective: adding $\mathbf{l}_s$ to the final-layer hidden state of the last token should increase the predictive likelihood of the next token in the corresponding language corpus. In this way, the learned vector captures language-specific distributional patterns and cultural-linguistic regularities, thereby anchoring generated responses within culturally informed reasoning frameworks without updating the base model.}

\subsection{Parametric Injection}

PSII injects identity information at two distinct levels and treats demographic and language vectors differently.

At the \textbf{prompt level}, agent profiles $P_i$ are included in the input prompt to provide semantic context. This ensures that the model has initial knowledge of the agent's demographic attributes before generation begins.

At the \textbf{representation level}, identity vectors are injected directly into the hidden states of the LLM. When predicting a response for a given sample, we first extract the demographic values $j_k$ for all variables $k$ and select the corresponding demographic vectors $\mathbf{d}_{k,j_k}$. During forward propagation, these vectors are injected into the hidden states of specific layers $\mathcal{L}_k$ using forward hooks. For token $t$ at layer $\mathcal{L}_k$, the hidden state is updated as:
\[
\tilde{h}^{(\mathcal{L}_k)}_t = h^{(\mathcal{L}_k)}_t + \mathbf{d}_{k,j_k} + \boldsymbol{\epsilon}_t,
\]
where $\boldsymbol{\epsilon}_t \sim \mathcal{N}(0, \sigma^2 \mathbf{I})$ is a small Gaussian noise vector used to induce intra-group heterogeneity. \textcolor{myred}{In practice, demographic vectors are constructed separately for each Transformer layer. At injection time, we select the vector corresponding to the target injection layer, so no vector is shared across layers.} \textcolor{myred}{During prompt encoding, the demographic vector is added to all prompt-token representations to establish a global demographic condition. During autoregressive generation, the same vector is added to the hidden representation of the newly generated token at each step, thereby continuously steering the response.}

To incorporate diverse cultural and linguistic context, we randomly select a language $s$ for each sample, translate the prompt into that language, and inject the corresponding language vector $\mathbf{l}_s$ at the \textbf{last layer} $L$. For token $t$ at the last layer during generation, the hidden state is updated as:
\[
\tilde{h}^{(L)}_t = h^{(L)}_t + \mathbf{l}_s.
\]
This mechanism introduces language-specific expression patterns and reasoning tendencies learned from the corpus, effectively providing a culturally informed anchor for model outputs.

\subsubsection{Noise Module}\label{sec:noisemethod_theory}

To model intra-group heterogeneity and prevent the over-essentialization of identity, we introduce a controlled Gaussian noise vector $\boldsymbol{\epsilon}_t$ when injecting demographic vectors. This noise adds small random perturbations to each demographic vector, simulating individual differences among agents with the same demographic attributes, thereby partially addressing the problem of insufficient internal diversity.

\[
\boldsymbol{\epsilon}_t \sim \mathcal{N}(0, \sigma^2 \mathbf{I}),
\]

where $\sigma$ is the standard deviation of the Gaussian noise, controlling the magnitude of perturbation applied to demographic vectors. Ideally, $\sigma$ should be calibrated to introduce variability without disrupting the model's core reasoning capabilities or demographic consistency. The specific calibration strategy and parameter selection are detailed in Section~\ref{sec:noise_calibration}.

\subsubsection{Layer-wise Hierarchical Injection}

Transformer-based LLMs capture information at different levels across their internal representations, ranging from linguistic constraints to abstract reasoning. Empirical studies in cognitive modeling and LLM behavior indicate that lower layers typically encode surface-level patterns, as well as syntactic and stylistic features; intermediate layers capture contextual information and background assumptions; and upper layers are responsible for abstract reasoning, value judgments, and final decision-making. Moreover, different personality traits and demographic attributes are processed differently across Transformer layers. Motivated by these observations, we adopt a hierarchical injection strategy, inserting demographic vectors into the layers that most naturally align with their semantic processing roles. Based on empirical analysis (see Section~\ref{sec:layer} and Appendix~\ref{sec:addres2.5} for details), demographic attributes are categorized into three hierarchical layer groups:

\textbf{Lower layers:} Govern behavioral feasibility and constraint processing. Demographic information related to responsibilities, life structure, or family obligations is injected at this level. Examples include: \textit{living with parents, primary income earner}.

\textbf{Intermediate layers:} Determine perspective and guide problem framing. Attributes reflecting experience sources, normative assumptions, or background context are injected at this level. Examples include: \textit{religion, immigration status}.

\textbf{Upper layers:} Govern final stance, value judgments, and decision level outputs. Attributes defining status, ideology, or structured social identity are injected at this level. Examples include: \textit{gender, marital status, education, employment, occupation, employer type, financial status, social class}.

\subsection{Theoretical Rationale}

The effectiveness of Parametric Social Identity Injection in maintaining stable demographic identities can be understood through the lens of representation-level control and the shortcomings of traditional persona modeling. Existing approaches typically rely on in-context learning (ICL) via textual prompts to induce individual differences. While prompt-based conditioning can guide the model at a surface level, it suffers from identity decay during long text generation or complex reasoning: as hidden representations propagate through multiple layers, the semantic impact of the prompt diminishes, leading to homogenized outputs and reduced diversity.

PSII mitigates this limitation by explicitly injecting demographic vectors into intermediate hidden states. These vectors act as structured, persistent constraints that continuously modulate token-level activations across layers, ensuring that each generated response adheres to the demographic identity of the synthetic agent. This mechanism fosters logical consistency and demographic stability, producing agents whose behavior is coherent, heterogeneous, and structured rather than merely repeating superficial prompt cues.

Additionally, the diversity of the injected vectors introduces controlled intra-group variability. By incorporating small, stochastic perturbations, PSII simulates natural heterogeneity within demographic categories, addressing the problem of identity essentialism and further enhancing the realism of the generated population. In effect, PSII provides both a stable anchor for identity and a flexible mechanism for capturing nuanced inter- and intra-group variation.

\section{Experimental Setup}

\subsection{Dataset and Evaluation Metrics}

We conduct experiments on the \textbf{World Values Survey (WVS)} dataset, a large-scale, cross-national survey designed to measure human values, beliefs, and socio-political attitudes across diverse cultural and demographic groups (\textcolor{myred}{See Appendix~\ref{app:dataset1} for details}). Following prior work, we use responses to \textbf{Q1--Q259} as target opinion questions, while the remaining questions \textbf{Q260--Q290} are reserved for demographic feature modeling and identity alignment.

\textcolor{myred}{For analysis, the 259 opinion questions are grouped into four high-level categories based on their thematic content: Personal Beliefs \& Life Outlook, Social Integration \& Perception, Political Engagement \& Institutional Identity, and Economic Development \& Progress. This regrouping is theory-driven and informed by established value-dimension frameworks, including the Inglehart--Welzel cultural map, as well as prior WVS-based simulation studies.} Details of the regrouping rationale are provided in Appendix~\ref{app:dataset2}.

Model performance is evaluated at the group level by comparing the distribution of model-generated responses with the human ground-truth distribution. We report \textbf{KL divergence} to measure distributional accuracy, where lower values indicate better alignment with human responses.  

To assess diversity, we compute \textbf{Entropy Deviation (ED)}, defined as the absolute difference between the normalized entropy of model-generated responses and that of human responses,
where a lower ED indicates closer diversity matching to the human distribution.

\subsection{Baseline Methods}

We compare our method against a diverse set of representative baselines:

\begin{itemize}
    \item \textbf{Direct}: Direct simulation without any fine-tuning or prompt engineering~\cite{santurkar2023whose}.
    
    \item \textbf{High-Temp}: High-temperature sampling to encourage output variability, with $\text{temperature}=2$~\cite{chung2023increasing}.
    
    \item \textbf{Multilingual}: Multilingual prompting to induce diversity by varying the language context~\cite{wang2025multilingual}.
    
    \item \textbf{DivReq}: Explicitly requesting diversity in the prompt, without additional structural constraints~\cite{wang2025multilingual}.
    
    \item \textbf{PE}: Prompt engineering with carefully designed persona templates to better approximate human respondents~\cite{yang2024large}.
    
    \item \textbf{SimVBG}: The SimVBG method~\cite{du-etal-2025-simvbg}, which constructs background stories and is guided by the Cognitive-Affective Personality System (CAPS) theory.

    \item \textbf{\textcolor{myred}{PV}}: \textcolor{myred}{Persona Vectors~\cite{chen2025persona}, a representation-level steering method that identifies persona-related directions in the model's activation space and uses them to control generated character traits.}
\end{itemize}

Detailed implementations of all baseline methods are described in Appendix~\ref{app:baseline}.

\subsection{Implementation Details}\label{sec:implede}

\begin{table*}
\caption{Main experimental results on the WVS dataset. We report KL divergence and Entropy Deviation (ED) for each method across four question categories and overall. Best-performing results are highlighted in bold.}
\label{tab:maintab}
\begin{tabular}{llcccccccccc}
\toprule
Model & Method & \multicolumn{2}{c}{Beliefs \& Life} & \multicolumn{2}{c}{Social Integration} & \multicolumn{2}{c}{Political Engagement} & \multicolumn{2}{c}{Economic Progress} & \multicolumn{2}{c}{Overall} \\
\cmidrule(lr){3-4} \cmidrule(lr){5-6} \cmidrule(lr){7-8} \cmidrule(lr){9-10} \cmidrule(lr){11-12}
 & & KL $\downarrow$ & ED $\downarrow$ & KL $\downarrow$ & ED $\downarrow$ & KL $\downarrow$ & ED $\downarrow$ & KL $\downarrow$ & ED $\downarrow$ & KL $\downarrow$ & ED $\downarrow$ \\
\midrule
\textbf{Qwen2.5-7B} & Direct & 1.5182 & 0.6664 & 0.8575 & 0.7406 & 2.2222 & 0.7625 & 1.6283 & 0.7932 & 1.3915 & 0.7340 \\
 & High-Temp & 1.1941 & 0.4754 & 0.7074 & 0.5981 & 1.9653 & 0.6266 & 1.1072 & 0.5272 & 1.1605 & 0.5778 \\
 & Multilingual & 1.0946 & 0.3706 & 0.5547 & 0.5112 & 1.9297 & 0.5000 & 1.1030 & 0.5371 & 1.0568 & 0.4811 \\
 & DivReq & 1.5442 & 0.6671 & 0.8383 & 0.7464 & 2.2166 & 0.7587 & 1.5282 & 0.7283 & 1.3813 & 0.7329 \\
 & PE & 1.4812 & 0.4050 & 0.7182 & 0.6292 & 1.8852 & 0.4847 & 1.5095 & 0.6046 & 1.2209 & 0.5443 \\
 & SimVBG & 0.8420 & 0.2494 & 0.3812 & 0.2942 & 1.0759 & 0.3025 & 1.0954 & 0.3646 & 0.6945 & 0.2908 \\
 & \textcolor{myred}{PV} & 1.5973 & 0.4796 & 0.7220 & 0.6551 & 1.8013 & 0.6563 & 0.9621 & 0.4417 & 1.1982 & 0.6102 \\
 & \cellcolor{yellow!30}\textbf{PSII} & \cellcolor{yellow!30}\textbf{0.6772} & \cellcolor{yellow!30}\textbf{0.0989} & \cellcolor{yellow!30}\textbf{0.1862} & \cellcolor{yellow!30}\textbf{0.0180} & \cellcolor{yellow!30}\textbf{0.9095} & \cellcolor{yellow!30}\textbf{0.0150} & \cellcolor{yellow!30}\textbf{0.3094} & \cellcolor{yellow!30}\textbf{0.0150} & \cellcolor{yellow!30}\textbf{0.4843} & \cellcolor{yellow!30}\textbf{0.0319} \\
\midrule
\textbf{Qwen2.5-14B} & Direct & 1.1020 & 0.7509 & 0.8343 & 0.7654 & 2.5925 & 0.7812 & 1.6882 & 0.8866 & 1.3982 & 0.7724 \\
 & High-Temp & 0.9247 & 0.6179 & 0.7293 & 0.6685 & 2.4167 & 0.7003 & 1.2345 & 0.6439 & 1.2435 & 0.6657 \\
 & Multilingual & 0.8584 & 0.5626 & 0.5423 & 0.5110 & 2.0181 & 0.5082 & 1.0812 & 0.6043 & 1.0258 & 0.5251 \\
 & DivReq & 1.2082 & 0.7718 & 0.8749 & 0.7751 & 2.6916 & 0.7556 & 1.7217 & 0.8545 & 1.4673 & 0.7730 \\
 & PE & 0.7451 & 0.4255 & 0.6418 & 0.5580 & 1.4125 & 0.4700 & 1.1097 & 0.4664 & 0.8905 & 0.5035 \\
 & SimVBG & \textbf{0.4102} & 0.2537 & 0.3562 & 0.2736 & 1.6102 & 0.3719 & 0.6910 & 0.2851 & 0.7215 & 0.2967 \\
 & \textcolor{myred}{PV} & 0.9186 & 0.6031 & 0.6655 & 0.6207 & 2.7785 & 0.6194 & 1.0715 & 0.5878 & 1.3005 & 0.6153 \\
 & \cellcolor{yellow!30}\textbf{PSII} & \cellcolor{yellow!30}0.5193 & \cellcolor{yellow!30}\textbf{0.1264} & \cellcolor{yellow!30}\textbf{0.3247} & \cellcolor{yellow!30}\textbf{0.2540} & \cellcolor{yellow!30}\textbf{1.1101} & \cellcolor{yellow!30}\textbf{0.2112} & \cellcolor{yellow!30}\textbf{0.4835} & \cellcolor{yellow!30}\textbf{0.1540} & \cellcolor{yellow!30}\textbf{0.5814} & \cellcolor{yellow!30}\textbf{0.2123} \\
\midrule
\textbf{Llama-3.1-8B} & Direct & 1.1033 & 0.6411 & 0.8232 & 0.6060 & 2.2543 & 0.6592 & 1.3212 & 0.7199 & 1.2856 & 0.6326 \\
 & High-Temp & 0.7738 & 0.3037 & 0.5194 & 0.3257 & 1.7162 & 0.3421 & 0.6542 & 0.3198 & 0.8970 & 0.3254 \\
 & Multilingual & 0.7657 & 0.4619 & 0.5298 & 0.3693 & 1.6478 & 0.3872 & 1.1913 & 0.6584 & 0.9071 & 0.4063 \\
 & DivReq & 1.0864 & 0.5541 & 0.9328 & 0.6202 & 1.8263 & 0.5825 & 1.2447 & 0.6700 & 1.2172 & 0.5992 \\
 & PE & 0.7124 & 0.3630 & 0.5219 & 0.4198 & 1.4478 & 0.3623 & 0.5560 & 0.2065 & 0.8095 & 0.3831 \\
 & SimVBG & 0.4275 & 0.2284 & 0.4879 & 0.2240 & 1.0170 & 0.3074 & 0.7253 & 0.2610 & 0.6298 & 0.2491 \\
 & \textcolor{myred}{PV} & 1.0344 & 0.5709 & 0.6095 & 0.4850 & 1.7505 & 0.5276 & 1.1692 & 0.6609 & 1.0263 & 0.5220 \\
 & \cellcolor{yellow!30}\textbf{PSII} & \cellcolor{yellow!30}\textbf{0.3776} & \cellcolor{yellow!30}\textbf{0.0157} & \cellcolor{yellow!30}\textbf{0.2305} & \cellcolor{yellow!30}\textbf{0.0254} & \cellcolor{yellow!30}\textbf{0.7494} & \cellcolor{yellow!30}\textbf{0.0386} & \cellcolor{yellow!30}\textbf{0.2911} & \cellcolor{yellow!30}\textbf{0.0601} & \cellcolor{yellow!30}\textbf{0.4017} & \cellcolor{yellow!30}\textbf{0.0040} \\
\midrule
\textbf{Mistral-24B} & Direct & 0.8621 & 0.4228 & 0.7749 & 0.5967 & 1.5736 & 0.5651 & 0.9845 & 0.4081 & 1.0158 & 0.5445 \\
 & High-Temp & 0.6893 & 0.1083 & 0.3788 & 0.0940 & 1.2904 & 0.2048 & 0.8349 & 0.2820 & 0.7064 & 0.1353\\
 & Multilingual & 0.6373 & 0.2698 & 0.4888 & 0.3609 & 1.4458 & 0.3030 & 0.6964 & 0.3057 & 0.7843 & 0.3246 \\
 & DivReq & 0.8306 & 0.1917 & 0.6905 & 0.3010 & 1.1870 & 0.3541 & 1.0052 & 0.2957 & 0.8662 & 0.2930 \\
 & PE & 0.6449 & 0.2925 & 0.6518 & 0.5587 & 1.3570 & 0.4041 & 1.0166 & 0.3885 & 0.8560 & 0.4558 \\
 & SimVBG & \textbf{0.3285} & 0.2228 & 0.3066 & 0.2458 & 1.5274 & 0.2364 & 0.6564 & 0.1738 & 0.6571 & 0.2353 \\
 & \textcolor{myred}{PV} & 0.7413 & 0.3592 & 0.6293 & 0.5850 & 1.7421 & 0.4808 & 0.7575 & 0.3323 & 0.9555 & 0.4999 \\
 & \cellcolor{yellow!30}\textbf{PSII} & \cellcolor{yellow!30}0.4795 & \cellcolor{yellow!30}\textbf{0.0013} & \cellcolor{yellow!30}\textbf{0.2278} & \cellcolor{yellow!30}\textbf{0.0880} & \cellcolor{yellow!30}\textbf{1.2505} & \cellcolor{yellow!30}\textbf{0.1244} & \cellcolor{yellow!30}\textbf{0.4134} & \cellcolor{yellow!30}\textbf{0.0252} & \cellcolor{yellow!30}\textbf{0.5607} & \cellcolor{yellow!30}\textbf{0.0774} \\
\bottomrule
\end{tabular}
\end{table*}

We evaluate all methods on four instruction-tuned large language models:
\textbf{Qwen2.5-7B-Instruct}, \textbf{Qwen2.5-14B-Instruct}, \textbf{Llama-3.1-8B-Instruct}, and \textbf{Mistral-24B-Instruct}.

From the full WVS dataset containing 97,220 respondents, we randomly sample 100 individuals to construct simulated agent populations for evaluation. \textcolor{myred}{Additional resampling experiments confirm that PSII is robust to random sampling variation, as shown in Appendix~\ref{app:sampling_robustness}.} Each simulated agent answers one question at a time, following the original WVS question order, and produces a numerical response consistent with the survey format.

To approximate value orientations, we train language-specific value vectors using the CulturaX dataset~\cite{nguyen2024culturax} for the five primary languages $s \in \{en, zh, ar, es, ru\}$ (\textcolor{myred}{See Appendix~\ref{app:dataset4}}), with training hyperparameters set to $n\_samples=20000$, $epochs=3$, and $learning\_rate=1\times10^{-3}$. 

Unless otherwise specified, generation uses default decoding parameters with temperature = $0.7$ and top\_k = $20$.

For methods involving stochastic identity perturbation, we inject Gaussian noise into the demographic vectors, with model-specific standard deviations: Qwen2.5-14B ($\sigma = 0.35$), Qwen2.5-7B ($\sigma = 0.30$), Mistral-24B ($\sigma = 0.09$), and Llama-3.1-8B ($\sigma = 0.07$).

Training of identity vectors is performed on a single \textbf{NVIDIA A100-SXM4-80GB} GPU, while inference can be completed on a single \textbf{NVIDIA A100-SXM4-40GB} GPU.

\subsection{Noise Calibration Strategy}\label{sec:noise_calibration}

To determine the optimal noise standard deviation $\sigma$ for each model, we propose a calibration metric termed \textbf{model sensitivity}. This metric quantifies the impact of noise on the predicted ranking of response options. 

Specifically, for each agent $i$ and question $j$, we compute the Mean Absolute Error (MAE) between the rankings of answer options predicted with and without noise, where the ranking reflects the position of each option among all candidate options for that specific question: 
\[
MAE_{i,j} = \frac{\left| \text{rank(answer}_{i,j}^\text{noise}) - \text{rank(answer}_{i,j}^\text{no noise}) \right|}{\text{number of options}}.
\]
The overall model sensitivity is computed by averaging over all sampled agents and questions. A higher sensitivity value indicates that the model's reasoning is easily disrupted by perturbations, requiring a smaller $\sigma$.

Empirically, we observed a linear correlation between the optimal noise level and model robustness. We specifically calibrate the noise standard deviation as:
\[
\sigma_\text{best} = \max(0, 0.4 - \text{model sensitivity}).
\]
Based on this calibration, if the model is too sensitive (sensitivity $> 0.4$), we set $\sigma=0$. The specific calibrated $\sigma$ values for each model used in our experiments are reported in Section~\ref{sec:implede}.

\section{Experiments and Results}
\subsection{Main Results}

\begin{figure*}[ht]
  \centering
  \includegraphics[width=0.8\linewidth]{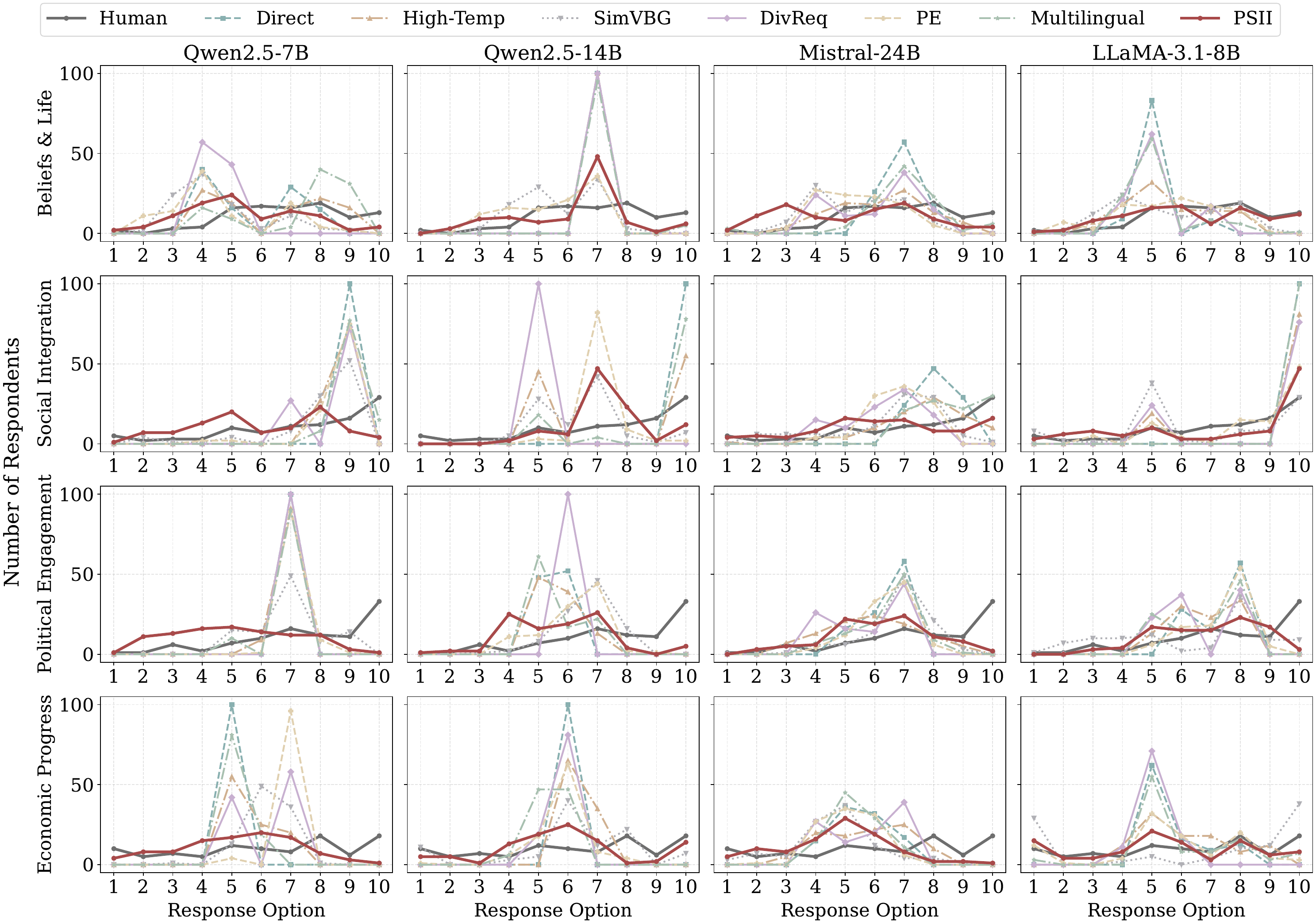}
  \caption{Response distributions for a randomly selected question from each of the four categories. \textcolor{myred}{PSII more closely matches the empirical response diversity observed in human survey data}, while baseline methods often concentrate on a few options.}
  \Description{Response distributions for a randomly selected question from each of the four categories. \textcolor{myred}{PSII more closely matches the empirical response diversity observed in human survey data}, while baseline methods often concentrate on a few options.}
  \label{fig:distribution}
\end{figure*}

We evaluate different simulation methods by measuring how well they reproduce human survey outcomes. The main results are summarized in Table~\ref{tab:maintab}. Overall, PSII consistently achieves the best trade-off between accuracy and diversity across models, parameter settings, and question categories, substantially narrowing the gap between simulated and human survey responses. \textcolor{myred}{See Appendix~\ref{sec:addres1} for additional results.}

Under direct prompting, Mistral-24B produces responses most similar to human data, followed by Llama-3.1-8B and then the Qwen series. After incorporating PSII, all models exhibit marked gains in both accuracy and diversity. Among them, Llama-3.1-8B benefits the most from PSII, followed by Qwen2.5-7B.

Performance also varies across question subsets. PSII performs best on the Economic Progress questions, while performance on Beliefs \& Life questions is relatively weaker. This suggests that questions related to economic evaluations may align more naturally with the structured representation shifts induced by PSII, while questions related to beliefs and values remain more challenging to simulate faithfully.

\textcolor{myred}{Compared to baseline methods, PSII reliably outperforms simple strategies such as Direct and Diversity Request. High-temperature decoding performs relatively well on Mistral-24B, but degrades substantially on other models, indicating that uncontrolled randomness alone does not provide a robust or general solution for realistic opinion simulation. Relative to Multilingual prompting and PE, PSII further introduces representation-level identity vector injection, resulting in better simulations. PSII also improves upon Persona Vectors by further incorporating value vectors, prompt-based profiles, controlled noise, and hierarchical layer-wise injection. Although SimVBG is a strong background-story-based baseline, PSII consistently achieves better overall performance across most settings.}

Finally, we observe a noticeable gap in ED achieved by PSII between Qwen2.5-7B and Qwen2.5-14B, which we attribute primarily to scale-dependent differences in internal representations. As PSII directly intervenes in hidden states, its impact on diversity is sensitive to model depth, hidden dimensionality, and representation geometry, which differ substantially across model scales. In practice, different scales also exhibit varying sensitivity to the injected noise, and the noise variance is therefore selected separately for each model to ensure stable generation, which may further contribute to the ED variation. Moreover, although the injection depth is aligned by relative position, differences in absolute depth imply that the injected layers may correspond to different functional stages in the two models, leading to specific diversity effects.

\subsection{Response Distribution Analysis}

\begin{table*}
\caption{Ablation study results on PSII across different models. Each row shows the impact of removing one component on accuracy and diversity metrics.}
\label{tab:ablation}
\begin{tabular}{lcccccccc}
\toprule
Setting & \multicolumn{2}{c}{Qwen2.5-7B} & \multicolumn{2}{c}{Qwen2.5-14B} & \multicolumn{2}{c}{Llama-3.1-8B} & \multicolumn{2}{c}{Mistral-24B} \\
\cmidrule(r){2-3} \cmidrule(r){4-5} \cmidrule(r){6-7} \cmidrule(r){8-9}
 & KL ↓ & ED ↓ & KL ↓ & ED ↓ & KL ↓ & ED ↓ & KL ↓ & ED ↓ \\
\midrule
\cellcolor{yellow!30}PSII (full model) & \cellcolor{yellow!30}\textbf{0.4843} & \cellcolor{yellow!30}0.0319 & \cellcolor{yellow!30}\textbf{0.5814} & \cellcolor{yellow!30}\textbf{0.2123} & \cellcolor{yellow!30}\textbf{0.4017} & \cellcolor{yellow!30}\textbf{0.0040} & \cellcolor{yellow!30}\textbf{0.5607} & \cellcolor{yellow!30}0.0774 \\
PSII w/o value vector & 0.5264 & 0.0348 & 0.6574 & 0.3040 & 0.4638 & 0.0416 & 0.6254 & 0.1723 \\
PSII w/o demographic vectors & 0.9705 & 0.3859 & 0.7893 & 0.3878 & 0.6681 & 0.2729 & 0.7185 & 0.3312 \\
PSII w/o prompt-based profile & 0.5024 & \textbf{0.0151} & 0.6688 & 0.2527 & 0.5575 & 0.0630 & 0.6149 & \textbf{0.0474} \\
PSII w/o parametric noise & 0.7237 & 0.2261 & 0.7332 & 0.3294 & 0.5673 & 0.2068 & 0.6813 & 0.2808 \\
PSII w/o layer-wise injection & 0.5313 & 0.0964 & 0.7665 & 0.2849 & 0.5225 & 0.1089 & 0.5780 & 0.0999 \\
\bottomrule
\end{tabular}
\end{table*}

To further analyze how different simulation methods capture behavioral variability, we examine the distributions of individual response choices, aiming to assess whether the response patterns generated by simulated agents reflect the dispersion observed in human survey data.

Specifically, we randomly select one question from each of the four question categories and visualize the response distributions produced by 100 simulated agents under different methods, alongside the corresponding empirical distributions from human respondents.

As shown in Figure~\ref{fig:distribution}, baseline methods tend to concentrate responses on a small number of options, exhibiting limited diversity. In contrast, PSII produces more evenly distributed responses that more closely match human data, indicating that it better preserves inter-agent heterogeneity at the level of concrete survey responses. These findings are consistent with our quantitative results and further validate the effectiveness of PSII in enhancing diversity.

\subsection{Ablation Studies}

To investigate the contribution of each component in PSII, we perform ablation studies in which we remove one module at a time, including the value vector, demographic vectors, prompt-based profile, parametric noise, and the layer-wise injection (modified to inject at 70\% of layers). The results are summarized in Table~\ref{tab:ablation} and report both simulation accuracy and diversity metrics across different models.

\textcolor{myred}{The results show that removing any single component reduces simulation accuracy, and most ablations also degrade diversity matching, indicating that all modules contribute meaningfully to PSII’s overall effectiveness.} Among them, the demographic vectors are the most critical for maintaining both accuracy and diversity, while parametric noise is especially important for enhancing diversity. Both the value vector and the layer-wise injection significantly contribute to gains in both accuracy and diversity. \textcolor{myred}{Prompt-based profiles act as semantic anchors, improving alignment to the target distribution but may constrain the output space, thus reducing diversity. This reflects a trade-off between accuracy and diversity. PSII combines prompt- and representation-level steering. While removing this module can possibly increase diversity, our ultimate goal is accurate distribution matching rather than maximizing diversity; therefore, retaining this module is the preferred design choice.}

Although the sensitivity to each component varies slightly across models, the overall trends are consistent, demonstrating that PSII is robust and effective across different LLM backbones.

These findings highlight that PSII’s performance arises from the complementary effects of its modules: structured identity and value representations provide coherent guidance for realistic responses, while carefully injected noise ensures sufficient behavioral variability, and layered interventions allow these effects to propagate effectively through the model.

%
%
%

\subsection{Layer-wise Injection Analysis}\label{sec:layer}

Before applying hierarchical injection in our full PSII framework, we conducted experiments on the Qwen2.5-7B model to determine the most suitable layers for injecting different personality attributes. Specifically, we tested all demographic attributes injected individually across layers 1 to 28, and recorded their impact on simulation performance, as shown in Appendix~\ref{sec:addres2}, Figure~\ref{fig:overall_layer}.

As illustrated, the optimal injection layer varies across different attributes. We selected the best layer for each attribute based on the layer that minimized the KL divergence, indicating the closest alignment with human response distributions. This approach ensures that each demographic attribute is incorporated at a point in the network where it most effectively influences the model’s output without disrupting stability.

Furthermore, we observed that when all vectors are injected simultaneously at a fixed layer, injecting at approximately 70\% of the network depth achieves the best overall performance. This finding motivated our choice in the ablation studies, where we conducted comparisons using injection at this 70\% layer as a representative baseline.

\section{Ethical Concerns and Societal Implications}

\textcolor{myred}{While PSII enables scalable and diversity-aware social simulation, its deployment requires careful consideration of several ethical and societal implications.}

\textcolor{myred}{A key concern involves data usage and privacy. Our experiments rely exclusively on publicly available human value datasets (e.g., World Values Survey), ensuring that no private or sensitive individual data is accessed or inferred. Researchers must continue to prioritize privacy-preserving practices when extending PSII to other datasets.}

\textcolor{myred}{Another risk lies in demographic modeling, which may introduce issues such as stereotyping or oversimplification. To mitigate these, we: (a) strictly follow established survey definitions; (b) focus on group-level (not individual-level) analysis; and (c) apply manual filtering. PSII is intended for controlled research settings only, not for real-world decision-making.}

\textcolor{myred}{Beyond modeling choices, the responsible use of synthetic agents is critical. Although LLM-based agents can simulate large-scale public opinion efficiently, unregulated deployment may lead to misuse, such as generating synthetic narratives for propaganda or social manipulation. Institutional and technical safeguards are therefore essential to prevent abuse, including access control, transparency reports, and usage auditing.}

\textcolor{myred}{Finally, context-aware application must be ensured. PSII-generated simulations should be interpreted as supplementary models rather than definitive reflections of societal opinions. Users are expected to consider limitations such as cultural nuances, minority representation, and model bias, and to carefully design experiments that mitigate potential harm or misinterpretation.}

\section{Conclusion}

\textcolor{myred}{This study investigates a key challenge faced by large language models} in public opinion simulation, specifically the phenomenon of "Diversity Collapse," where synthetic populations exhibit significant inter-group homogenization and insufficient intra-group representativeness. By analyzing the internal representation dynamics, we reveal that this issue stems, in part, from a systematic contraction of representations in the upper layers of the Transformer, causing distinct social identities to converge toward a uniform state at the end of the reasoning chain. To address this, we propose the Parametric Social Identity Injection (PSII) framework.
By directly embedding parametric vectors of demographic attributes and value orientations into the intermediate hidden states, PSII achieves stable guidance and fine-grained modulation of identity attributes at the representation level. \textcolor{myred}{This allows identity signals to persist throughout generation and enables structured, controllable diversity aligned with population attributes.}
Extensive experiments on the World Values Survey (WVS) demonstrate that PSII consistently enhances the distributional fidelity and diversity of simulation results across multiple mainstream open-source LLMs, significantly outperforming existing baseline methods. Our work not only elucidates the impact of model representations on simulation quality but also provides a \textcolor{myred}{practical} technical pathway for constructing large-scale, high-fidelity, and diversity-aware digital twin social simulations.

\begin{acks}
\textcolor{myred}{This work is supported by the Research Project of Quancheng Laboratory, China (Grant No.~QCL20250105), the National Natural Science Foundation of China No.~62502260, and the Postdoctoral Fellowship Program of China Postdoctoral Science Foundation No.~GZC20240833.}
\end{acks}

\clearpage
\bibliographystyle{ACM-Reference-Format}
\balance
\bibliography{sample-base}

\clearpage
\appendix

\section{Limitations}

Despite its advantages, PSII has several limitations that warrant attention.

\textbf{Dependence on demographic coverage.} The quality of synthetic populations is constrained by the granularity and completeness of available demographic data. Rare or underrepresented groups may still be insufficiently modeled, potentially limiting simulation fidelity in these populations.

\textbf{Scope of identity modeling.} Current PSII vectors primarily encode coarse-grained demographic attributes and value orientations. Fine-grained personality traits, dynamic opinion changes, or context-specific behavioral nuances are not explicitly captured and may require additional mechanisms.

\section{Baseline Implementation Details}\label{app:baseline}

We select baseline methods from prior work that are applicable to the World Values Survey (WVS) setting. Specifically, we focus on approaches designed for population-level simulation and structured or semi-structured survey settings. Several diversity-enhancing methods proposed in the literature primarily target semantic or stylistic diversity in open-ended text generation, which are not suitable for closed-form, fixed-choice survey questions such as those in WVS. These methods are therefore excluded from our comparison.

Below, we describe the implementation details of each baseline considered in our experiments.

\tcbset{
    myboxstyle/.style={
        colback=gray!10,          
        colframe=black!50,        
        coltitle=white,           
        colbacktitle=gray!70,     
        fonttitle=\bfseries,      
        rounded corners,          
        boxrule=0.5mm,            
        top=1mm, bottom=1mm,
        left=2mm, right=2mm
    }
}

\newtcolorbox{mybox}[2][]{  
    myboxstyle,
    title=#2,
    #1                       
}

\textbf{Direct.}
The Direct baseline performs LLM-based simulation without any explicit mechanisms for diversity control or identity conditioning. For each survey question, the model is prompted to generate a response directly, serving as a minimal and commonly used reference setting in prior social simulation work.

\begin{mybox}{Direct Prompt}
\small\sffamily
Forget you are an AI model. Simulate a human being.
\end{mybox}

\textbf{High-Temp.}
The High-Temp baseline applies high-temperature sampling to encourage output variability. It uses the same prompting strategy as Direct, but sets the sampling temperature to 2. This baseline tests whether stochastic decoding alone is sufficient to induce population-level diversity.

\textbf{Multilingual.}
The Multilingual baseline aims to increase diversity through linguistic variation. The original prompt used in Direct is translated into five languages: Arabic (ar), English (en), Spanish (es), Russian (ru), and Chinese (zh). For each simulated individual, one language is randomly selected using a fixed random seed (42). Apart from the prompt language, all other settings are identical to the Direct baseline.

\textbf{DivReq.}
The DivReq baseline introduces an explicit diversity request at the prompt level. Specifically, we augment the Direct prompt with an additional instruction encouraging output diversity. This baseline assesses whether prompt-level diversity requests alone are sufficient to alleviate output homogenization.

\begin{mybox}{Additional Instruction}
\small\sffamily
Please try to be as diverse as possible.
\end{mybox}

\textbf{PE (Prompt Engineering).}
The PE baseline conditions the model on structured demographic profiles constructed from real human samples in the WVS dataset.
\textcolor{myred}{Specifically, we convert the self-reported demographic information of each subject in the dataset (e.g., age, gender, income level, education, religious beliefs, etc.) into natural language descriptions, which are then used as prompts to input into a LLM, thereby guiding the model to simulate the responses, attitudes, or behavioral reactions of that individual.}

\begin{mybox}{PE Prompt}
\small\sffamily
Please answer based on the following personal profile:

You are a \{sex\}. You are aged \{age\}. You live in \{country\}. You are \{citizenship\}. \{immigrant\_self\_status\}. \{immigrant\_mother\_status\}. \{immigrant\_father\_status\}. You live in a \{urban\_rural\} area, \{settlement\_type\}, with a population of \{settlement\_size\}. Your household consists of \{household\_size\} members. You \{live\_with\_parents\}. You speak \{home\_language\} at home. You are \{marital\_status\}, have \{num\_children\} children. Your education level is \{education\_self\}. Your spouse's education level is \{education\_spouse\}. Your father's education level is \{education\_father\}. Your mother's education level is \{education\_mother\}. You are \{employment\_self\}. You \{occupation\_self\}. You work in the \{work\_sector\}. Your spouse is \{employment\_spouse\}. Your spouse \{occupation\_spouse\}. Your father \{occupation\_father14\} when you were 14 years old. You \{chief\_wage\_earner\}. During the past year, your family \{family\_savings\}. You belong to the \{social\_class\}. Your income is in income group \{income\_decile\} (1 = lowest, 10 = highest). \{religion\_status\}. Your ethnic group is \{ethnic\_group\}.
\end{mybox}

\begin{mybox}{PE Example}
\small\sffamily
Please answer based on the following personal profile:

You are a female. You are aged 35. You live in Chile. You are a citizen of this country. You are born locally. Your mother is born locally. Your father is born locally. You live in a urban area, regional center, with a population of 100,000-500,000. Your household consists of 2 members. You do not live with parents. You speak Spanish; Castilian at home. You are living together as married, have 0 children. Your education level is Master's degree. Your spouse's education level is Master's degree. Your father's education level is upper secondary education. Your mother's education level is upper secondary education. You are employed full-time. You work in professional/technical fields. You work in the private business/industry. Your spouse is employed full-time. Your spouse works in professional/technical fields. Your father worked as a semi-skilled worker when you were 14 years old. You are not the chief wage earner. During the past year, your family had spent some savings. You belong to the lower middle class. Your income is in income group group 5 (1 = lowest, 10 = highest). You identify as Roman Catholic. Your ethnic group is White, Caucasian.
\end{mybox}

This baseline represents a strong prompt-based identity conditioning approach commonly used in prior social simulation work. Similarly, in PSII, the prompt-level injection of demographic information is implemented in the same manner.

\textbf{SimVBG.}
SimVBG first converts structured demographic profiles into a coherent background narrative. Guided by the Cognitive-Affective Personality System (CAPS) theory, it then generates candidate responses independently along three dimensions: cognitive, affective, and behavioral, and aggregates them to produce the final simulated response. We follow the original paper’s implementation and parameter settings~\cite{du-etal-2025-simvbg}.

\textcolor{myred}{\textbf{PV.}}
\textcolor{myred}{Persona Vectors~\cite{chen2025persona}, which steer LLM behavior by identifying persona-related directions in the activation space. While the original PV method mainly targets general behavioral traits, such as ``evil'', our task focuses on social survey simulation. We therefore adapt PV using the same vector-construction and steering procedure to build demographic steering vectors from demographic descriptions. Following the reported optimal configuration, we apply PV at 70\% of the network depth with response-level steering and coefficient 2.}

\section{Dataset Details}

\subsection{World Values Survey Dataset}\label{app:dataset}

\subsubsection{\textcolor{myred}{Dataset Overview}}\label{app:dataset1}

The \textbf{World Values Survey (WVS)} is a large-scale, cross-national survey that investigates human values, beliefs, and social attitudes across countries and time. It covers a broad range of topics related to individual life outlooks, social norms, political orientations, economic values, and demographic characteristics. Due to its wide thematic coverage and standardized questionnaire design, WVS has become a widely used benchmark dataset in the social sciences for studying cultural variation and population-level heterogeneity.

The original WVS questionnaire organizes questions into multiple thematic categories, each corresponding to a contiguous range of question identifiers. Table~\ref{tab:wvs_original_categories} summarizes the original value categories along with their corresponding question IDs.

\begin{table}[H]
\centering
\caption{Original value categories and question mappings in the World Values Survey (WVS).}
\label{tab:wvs_original_categories}
\resizebox{\linewidth}{!}{
\begin{tabular}{ll}
\toprule
\textbf{Original WVS Category} & \textbf{Question IDs} \\
\midrule
Social Values, Attitudes \& Stereotypes & Q1--Q45 \\
Happiness and Well-Being & Q46--Q56 \\
Social Capital, Trust \& Organizational Membership & Q57--Q105 \\
Economic Values & Q106--Q111 \\
Corruption & Q112--Q120 \\
Migration & Q121--Q130 \\
Security & Q131--Q151 \\
Postmaterialist Index & Q152--Q157 \\
Science and Technology & Q158--Q163 \\
Religious Values & Q164--Q175 \\
Ethical Values and Norms & Q176--Q198 \\
Political Interest and Participation & Q199--Q234 \\
Political Culture and Political Regimes & Q235--Q259 \\
Demographics & Q260--Q290 \\
\bottomrule
\end{tabular}
}
\end{table}

\subsubsection{\textcolor{myred}{Reorganized Question Groups}}\label{app:dataset2}

For the purpose of population simulation and value modeling, we reorganize the original WVS categories into four higher-level semantic groups that better reflect underlying dimensions of human values and social cognition. This reclassification is guided by conceptual coherence and prior theoretical work in sociology and political science, rather than the original questionnaire ordering. The four categories are described below.

\begin{itemize}
    \item \textbf{Personal Beliefs and Life Outlook:} Includes questions on \textbf{happiness and well-being, religious values, ethical values, and the postmaterialism index}. These items capture individuals’ internal belief systems, moral boundaries, and subjective evaluations of life quality, representing the most personal and deeply rooted dimensions of human values.
    
    \item \textbf{Social Integration and Perception:} Includes \textbf{social values, norms, and stereotypes; social capital, trust, and organizational membership; and perceptions of migration and security}. These questions focus on how individuals relate to others, perceive social cohesion, and evaluate out-groups and societal stability, reflecting levels of social integration and interpersonal trust.
    
    \item \textbf{Political Engagement and Institutional Identity:} Includes \textbf{political interest and participation, political culture and regime attitudes, and perceptions of corruption}. This category emphasizes the relationship between individuals and political institutions, capturing civic engagement, regime legitimacy, and evaluations of governance quality.
    
    \item \textbf{Economic Development and Progress:} Includes \textbf{economic values and perceptions of science and technology}. This category reflects attitudes toward resource allocation, economic organization, and technological progress, representing society’s orientation toward material development and future growth.
\end{itemize}

\subsubsection{\textcolor{myred}{Demographic Features for Identity Modeling}}\label{app:dataset3}

In addition to value-related questions, the WVS provides rich demographic information, which we leverage for identity modeling. Specifically, we use the full set of questions Q260–Q290 to construct profile descriptions, while a subset of these attributes is employed to build demographic vectors (see Table~\ref{tab:demographic_vectors}). \textcolor{myred}{The selection follows three principles. First, we prioritize \textit{cross-survey availability}, selecting variables that are commonly collected in major social surveys, such as the European Social Survey (ESS), Chinese General Social Survey (CGSS), International Social Survey Programme (ISSP), and Comparative Study of Electoral Systems (CSES). Second, we emphasize \textit{structural explanatory power}. Variables such as age, gender, education, income, employment, marital status, religion, and household composition capture fundamental social positions and are widely used to explain behavioral and attitudinal differences in survey research. Third, we prefer attributes with \textit{relative stability}. Compared with issue-specific opinions or transient attitudes, these demographic characteristics are more stable and therefore more suitable as underlying identity conditions for synthetic agents.}

\begin{table}[H]
\centering
\caption{Demographic attributes used for constructing demographic vectors.}
\label{tab:demographic_vectors}
\resizebox{\linewidth}{!}{
\begin{tabular}{ll}
\toprule
\textbf{Question ID} & \textbf{Question Topic} \\
\midrule
Q260 & Respondent’s sex \\
Q263 & Native-born or immigrant status \\
Q271 & Living with parents or parents-in-law \\
Q273 & Marital status \\
Q275 & Highest completed education level \\
Q279 & Employment status and working hours \\
Q281 & Occupational group \\
Q284 & Type of employer (government, private, nonprofit) \\
Q285 & Chief wage earner of the household \\
Q286 & Family financial situation in the past year \\
Q287 & Self-identified social class \\
Q288 & Household income decile (1--10 scale) \\
Q289 & Religious denomination \\
\bottomrule
\end{tabular}
}
\end{table}

\subsubsection{\textcolor{myred}{Language Distribution for Value Vector Training}}\label{app:dataset4}

\begin{figure}[H]
  \centering
  \includegraphics[width=0.7\linewidth]{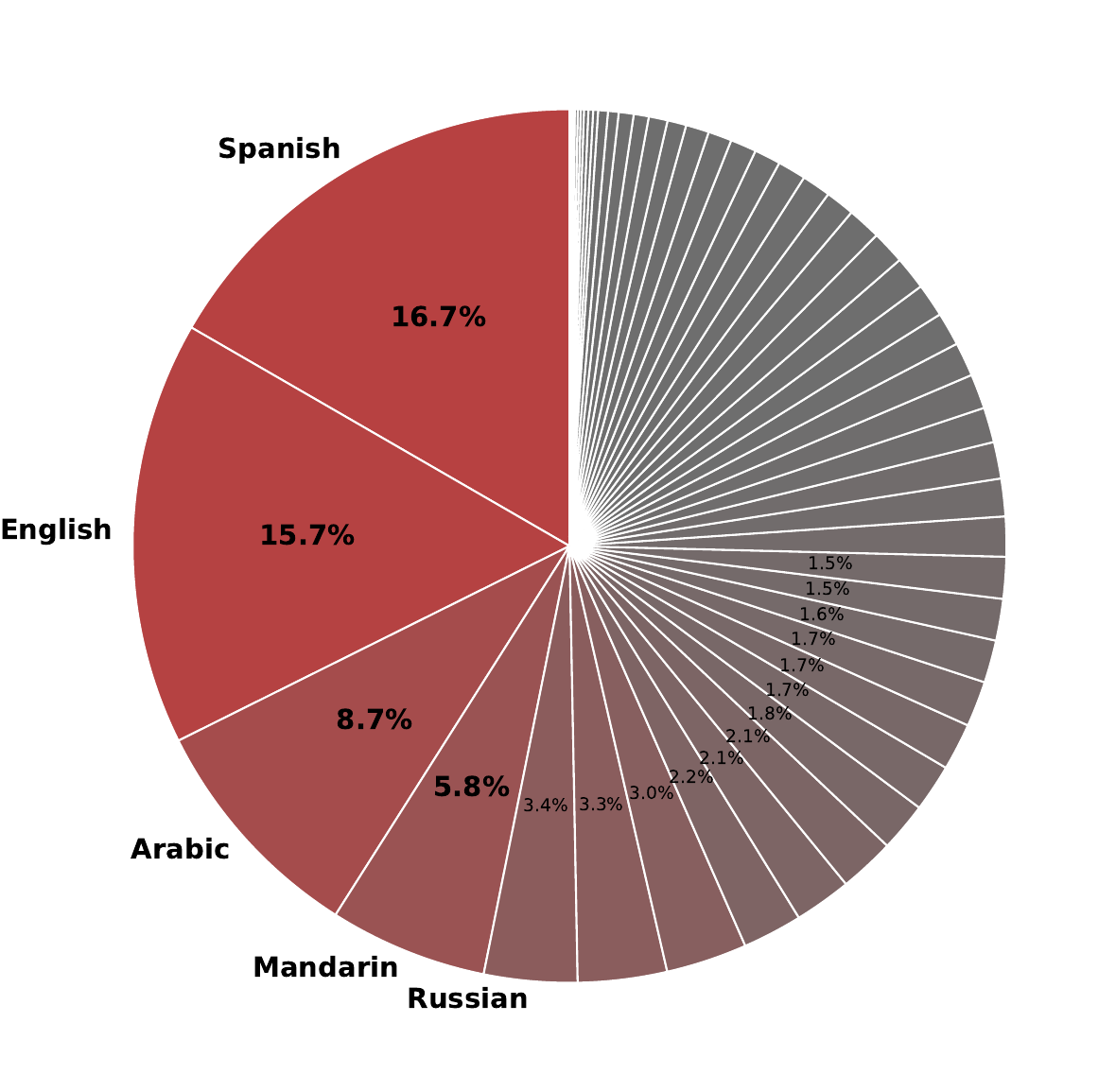}
  \caption{Language distribution in the WVS dataset.}
  \Description{Language distribution in the WVS dataset.}
  \label{fig:wvs_language}
\end{figure}

Figure~\ref{fig:wvs_language} shows the distribution of languages used by respondents in Wave 7 of WVS. For the purpose of constructing value vectors, we selected the five most common languages as training inputs. This choice ensures that the model captures the major linguistic groups in the dataset while maintaining computational efficiency, and allows the resulting value vectors to reflect the heterogeneity associated with language-based identity.

\subsection{Demographic Vectors Dataset}
\label{app:dv_dataset}

Before constructing the demographic vectors, we first generate a set of survey questions and persona instructions for each demographic feature. Specifically, for each demographic attribute, we design 40 social survey questions that are highly relevant to that feature. Then, for each possible value of the attribute (e.g., "Male" vs. "Female" for the gender feature), we generate 5 distinct persona instructions. Together, the questions and persona instructions form the \emph{Demographic Vectors} used in our experiments.

All questions and instructions are generated using \textbf{GPT-4o}. The prompts used to generate the dataset are as follows:

\begin{mybox}{Generate Question Prompt}
\footnotesize\sffamily
You are tasked with designing 40 social survey questions for a specific demographic \textbf{Feature Category}.

The target Feature Category is:\{FEATURE\_CATEGORY\}

Design 40 social survey questions that are \textbf{highly relevant} to the \textless feature\_category\textgreater.
These questions should elicit responses that naturally reflect the attitudes, concerns, and potential biases related to this category (e.g., questions about financial policy for "Income", or questions about work-life balance for "Employment Status"). These questions will be used for all values within this category (e.g., Low Income, Middle Income, High Income).

Organize your response in the following JSON format:

\textless output\_format\textgreater

\{

  "feature\_category": "The category used (e.g., Income)",
  
  "questions": [
  
    "question 1",
    
    "question 2",
    
    ...,
    
    "question 40"
    
  ]
  
\}

\textless/output\_format\textgreater

Your final output must only include the JSON object.

\end{mybox}

\begin{mybox}{Generate Persona Instructions Prompt}
\footnotesize\sffamily
You are tasked with creating immersive persona instructions for a social survey simulation. You will be given a Feature Category and a specific Feature Value.

Generate a list of five distinct \textbf{System/Persona Instructions} that command the model to adopt the identity, background, and typical attitudes associated with the Feature Value. These instructions will be used as the model's system message before it answers the survey questions.

The Feature Category is: \{FEATURE\_CATEGORY\} 

The specific Feature Value is: \{FEATURE\_VALUE\}

Create 5 distinct persona instructions. Ensure each instruction is unique in its framing, tone, or specific situational context.

Example for FEATURE\_CATEGORY: "Income", FEATURE\_VALUE: "Low Income":

\textless example\_instruction\textgreater

"Your name is Alex. You are currently struggling financially, working two minimum-wage jobs just to cover rent and basic necessities. Your outlook on economic policy is cautious and skeptical of large corporations. Answer all questions from this perspective."

\textless/example\_instruction\textgreater

Organize your response in the following JSON format:

\textless output\_format\textgreater

\{

  "feature\_category": "\{FEATURE\_CATEGORY\}",
  
  "feature\_value": "\{FEATURE\_VALUE\}",
  
  "instructions": [
  
    "persona instruction 1",
    
    "persona instruction 2",
    
    ...,
    
    "persona instruction 5"
    
  ]
  
\}

\textless/output\_format\textgreater

Your final output must only include the JSON object.
\end{mybox}

\section{\textcolor{myred}{Robustness Checks}}

\subsection{\textcolor{myred}{Robustness of Demographic Vector Construction}}
\label{app:vector_robustness}

\begin{table*}
\centering
\small
\caption{Average pairwise semantic similarity of demographic descriptions within each vector-construction setting. Similarity is computed using paraphrase-multilingual-MiniLM-L12-v2.}
\label{tab:semantic_consistency}
\resizebox{\linewidth}{!}{
\begin{tabular}{lccccccccccc}
\toprule
Setting & Claude & DeepSeek & GPT-4o & 4o-V1 & 4o-V2 & 4o-V3 & 4o-V4 & 4o-S1 & 4o-S123 & 4o-S42 & GPT-5-mini \\
\midrule
Avg. similarity 
& 0.8742 & 0.9337 & 0.9341 & 0.9353 & 0.9378 & 0.9196 & 0.9304 & 0.9357 & 0.9387 & 0.9397 & 0.9043 \\
\bottomrule
\end{tabular}
}
\end{table*}

\textcolor{myred}{Demographic vectors in PSII are constructed from LLM-generated attribute descriptions. This process may potentially introduce sensitivity to the choice of instruction model, prompt template, or random seed. To examine whether the effectiveness of PSII depends on a specific vector-construction setting, we conduct additional robustness analyses on Qwen2.5-7B by reconstructing demographic vectors under different settings and evaluating their downstream simulation performance.}

\subsubsection{\textcolor{myred}{Sensitivity to Instruction Models and Random Seeds}}

\textcolor{myred}{We first evaluate the downstream robustness of PSII when demographic vectors are constructed using different instruction-model settings. Specifically, we compare the original GPT-4o setting, GPT-4o with three random seeds, and GPT-5-mini. For each setting, we reconstruct the demographic vectors and evaluate PSII on Qwen2.5-7B using the same WVS evaluation protocol as in the main experiments.}

\textcolor{myred}{As shown in Table~\ref{tab:vector_construction_sensitivity}, the downstream performance remains stable across construction settings. The average KL divergence is 0.5017 and the average ED is 0.0236. The variances are small for both metrics, $2.64\times 10^{-4}$ for KL and $5.5\times 10^{-5}$ for ED, indicating that PSII is not highly sensitive to a particular instruction-model instance or random seed used for demographic-vector construction.}

\begin{table}[H]
\centering
\small
\caption{Sensitivity analysis of demographic-vector construction on Qwen2.5-7B. We reconstruct demographic vectors under different instruction-model and seed settings and report downstream KL divergence and Entropy Deviation (ED).}
\label{tab:vector_construction_sensitivity}
\begin{tabular}{lcc}
\toprule
Setting & KL $\downarrow$ & ED $\downarrow$ \\
\midrule
GPT-4o, default setting & 0.4843 & 0.0319 \\
GPT-4o, seed = 1 & 0.5221 & 0.0294 \\
GPT-4o, seed = 123 & 0.5142 & 0.0245 \\
GPT-4o, seed = 42 & 0.4884 & 0.0173 \\
GPT-5-mini, default setting & 0.4997 & 0.0148 \\
\midrule
Average & 0.5017 & 0.0236 \\
Variance & 0.000264 & 0.000055 \\
\bottomrule
\end{tabular}
\end{table}

\subsubsection{\textcolor{myred}{Semantic Consistency Across Construction Settings}}

\textcolor{myred}{We consider 11 settings, including four instruction models, four GPT-4o prompt-template variants, and three GPT-4o random seeds. Specifically, the compared settings include GPT-4o, GPT-5-mini, DeepSeek-V3, Claude-Haiku-4.5, four GPT-4o prompt variants, and GPT-4o with seeds 1, 123, and 42. For each setting, we encode the generated demographic descriptions using paraphrase-multilingual-MiniLM-L12-v2 and compute the average pairwise semantic similarity among all descriptions generated under that setting. As shown in Table~\ref{tab:semantic_consistency}, most settings achieve an average pairwise similarity above 0.90. Claude-Haiku-4.5 obtains a slightly lower but still high similarity score of 0.8742. These results suggest that the demographic semantics used for vector construction are largely consistent across model, prompt, and seed variations.}

\subsubsection{\textcolor{myred}{Manual Filtering of Generated Descriptions}}

\textcolor{myred}{To further reduce spurious or biased attribute descriptions, we manually inspect the generated demographic descriptions before vector computation. Entries containing explicit stereotypes, offensive expressions, or content unrelated to the target demographic attribute are removed. This filtering step is used only to improve the quality of demographic-vector construction and does not modify the WVS ground-truth responses, evaluation labels, or any downstream evaluation data.}

\textcolor{myred}{Together, these analyses indicate that the demographic-vector construction process is robust to moderate variations in instruction models, prompt templates, and random seeds, and that the constructed vectors preserve consistent demographic semantics across settings.}

\subsection{\textcolor{myred}{Sampling Robustness}}
\label{app:sampling_robustness}

\textcolor{myred}{In the main experiments, we randomly sample 100 respondents from the full WVS dataset to construct the simulated population. This choice follows prior work such as SimVBG and balances computational cost with comparability to existing baselines. However, because the full WVS dataset contains 97,220 respondents, it is important to examine whether the results are sensitive to random sampling variation.}

\textcolor{myred}{To evaluate sampling robustness, we independently sample five groups of respondents, each containing 100 individuals, and evaluate PSII on Qwen2.5-7B using the same experimental protocol as in the main experiments. As shown in Table~\ref{tab:sampling_robustness}, PSII achieves highly consistent performance across different samples. The average KL divergence is 0.4732 and the average ED is 0.0290. The variances are small for both metrics, $6.39\times 10^{-4}$ for KL and $1.5\times 10^{-5}$ for ED, indicating that the performance of PSII is robust to random variation in respondent sampling.}

\begin{table}[H]
\centering
\small
\caption{Sampling robustness analysis on Qwen2.5-7B. We independently sample five groups of 100 respondents from the WVS dataset and report the overall KL divergence and Entropy Deviation (ED).}
\label{tab:sampling_robustness}
\begin{tabular}{lcc}
\toprule
Sample group & KL $\downarrow$ & ED $\downarrow$ \\
\midrule
Group A: original sample & 0.4843 & 0.0319 \\
Group B: resample 1 & 0.4742 & 0.0302 \\
Group C: resample 2 & 0.4660 & 0.0231 \\
Group D: resample 3 & 0.4362 & 0.0272 \\
Group E: resample 4 & 0.5051 & 0.0324 \\
\midrule
Average & 0.4732 & 0.0290 \\
Variance & 0.000639 & 0.000015 \\
\bottomrule
\end{tabular}
\end{table}

\section{Additional Experimental Results}

In this section, we present additional results to complement the main experiments. 

\begin{table*}
\caption{Main experimental results on the WVS dataset. We report JS divergence and MAE for each method across four question categories and overall. Best-performing results are highlighted in bold.}
\label{maintab2}
\begin{tabular}{llcccccccccc}
\toprule
Model & Method & \multicolumn{2}{c}{Beliefs \& Life} & \multicolumn{2}{c}{Social Integration} & \multicolumn{2}{c}{Political Engagement} & \multicolumn{2}{c}{Economic Progress} & \multicolumn{2}{c}{Overall} \\
\cmidrule(lr){3-4} \cmidrule(lr){5-6} \cmidrule(lr){7-8} \cmidrule(lr){9-10} \cmidrule(lr){11-12}
 & & JS $\downarrow$ & MAE $\downarrow$ & JS $\downarrow$ & MAE $\downarrow$ & JS $\downarrow$ & MAE $\downarrow$ & JS $\downarrow$ & MAE $\downarrow$ & JS $\downarrow$ & MAE $\downarrow$ \\
\midrule
\textbf{Qwen2.5-7B} & Direct & 0.6822 & 0.2264 & 0.5625 & 0.3157 & 0.7013 & 0.3067 & 0.7611 & 0.1671 & 0.6330 & 0.2884 \\
 & High-Temp & 0.6167 & 0.2012 & 0.5116 & 0.2836 & 0.6356 & 0.2731 & 0.6460 & 0.1370 & 0.5722 & 0.2574 \\
 & Multilingual & 0.5350 & 0.1747 & 0.4487 & 0.2488 & 0.5681 & 0.2341 & 0.6421 & 0.1373 & 0.5070 & 0.2247 \\
 & DivReq & 0.7007 & 0.2375 & 0.5632 & 0.3167 & 0.6908 & 0.3005 & 0.7494 & 0.1649 & 0.6337 & 0.2893 \\
 & PE & 0.5909 & 0.1799 & 0.5068 & 0.2861 & 0.5463 & 0.2224 & 0.7050 & 0.1524 & 0.5435 & 0.2414 \\
 & SimVBG & 0.4991 & 0.1512 & 0.3436 & 0.1845 & 0.4281 & 0.1620 & 0.5877 & 0.1184 & 0.4090 & 0.1687 \\
 & \textcolor{myred}{PV} & 0.6013 & 0.2092 & 0.5166 & 0.2943 & 0.6367 & 0.2818 & 0.6012 & 0.1296 & 0.5697 & 0.2662 \\
 & \cellcolor{yellow!30}\textbf{PSII} & \cellcolor{yellow!30}\textbf{0.3409} & \cellcolor{yellow!30}\textbf{0.1017} & \cellcolor{yellow!30}\textbf{0.2174} & \cellcolor{yellow!30}\textbf{0.1481} & \cellcolor{yellow!30}\textbf{0.2892} & \cellcolor{yellow!30}\textbf{0.1226} & \cellcolor{yellow!30}\textbf{0.2922} & \cellcolor{yellow!30}\textbf{0.0576} & \cellcolor{yellow!30}\textbf{0.2650} & \cellcolor{yellow!30}\textbf{0.1277} \\
\midrule
\textbf{Qwen2.5-14B} & Direct & 0.6269 & 0.2313 & 0.5587 & 0.3102 & 0.6826 & 0.2812 & 0.7697 & 0.1700 & 0.6154 & 0.2800 \\
 & High-Temp & 0.5838 & 0.2081 & 0.5188 & 0.2837 & 0.6437 & 0.2605 & 0.6873 & 0.1488 & 0.5731 & 0.2560 \\
 & Multilingual & 0.5501 & 0.1910 & 0.4432 & 0.2391 & 0.5545 & 0.2186 & 0.6500 & 0.1375 & 0.5041 & 0.2192 \\
 & DivReq & 0.6485 & 0.2314 & 0.5692 & 0.3172 & 0.6960 & 0.2937 & 0.7737 & 0.1700 & 0.6286 & 0.2868 \\
 & PE & 0.4904 & 0.1598 & 0.4776 & 0.2681 & 0.5237 & 0.2080 & 0.6183 & 0.1263 & 0.4990 & 0.2236 \\
 & SimVBG & 0.3615 & \textbf{0.1136} & 0.3407 & 0.1903 & 0.4755 & 0.1956 & 0.4832 & 0.1055 & 0.3879 & 0.1724 \\
 & \textcolor{myred}{PV} & 0.5675 & 0.2053 & 0.4932 & 0.2743 & 0.6142 & 0.2487 & 0.6386 & 0.1378 & 0.5473 & 0.2472 \\
 & \cellcolor{yellow!30}\textbf{PSII} & \cellcolor{yellow!30}\textbf{0.3557} & \cellcolor{yellow!30}0.1145 & \cellcolor{yellow!30}\textbf{0.3217} & \cellcolor{yellow!30}\textbf{0.1846} & \cellcolor{yellow!30}\textbf{0.3838} & \cellcolor{yellow!30}\textbf{0.1527} & \cellcolor{yellow!30}\textbf{0.4049} & \cellcolor{yellow!30}\textbf{0.0855} & \cellcolor{yellow!30}\textbf{0.3491} & \cellcolor{yellow!30}\textbf{0.1573} \\
\midrule
\textbf{Llama-3.1-8B} & Direct & 0.6156 & 0.2299 & 0.5421 & 0.3084 & 0.6508 & 0.2781 & 0.7000 & 0.1495 & 0.5933 & 0.2771 \\
 & High-Temp & 0.4860 & 0.1706 & 0.4218 & 0.2324 & 0.5108 & 0.2052 & 0.5215 & 0.1018 & 0.4632 & 0.2066 \\
 & Multilingual & 0.5105 & 0.1854 & 0.4241 & 0.2442 & 0.5042 & 0.2004 & 0.6418 & 0.1404 & 0.4731 & 0.2158 \\
 & DivReq & 0.6069 & 0.2247 & 0.5751 & 0.3386 & 0.6229 & 0.2620 & 0.6869 & 0.1484 & 0.5995 & 0.2863 \\
 & PE & 0.4689 & 0.1568 & 0.4249 & 0.2343 & 0.4868 & 0.1960 & 0.4087 & 0.0791 & 0.4495 & 0.2012 \\
 & SimVBG & 0.3539 & 0.1255 & 0.3612 & 0.2131 & 0.4339 & 0.1764 & 0.4624 & 0.1157 & 0.3841 & 0.1811 \\
 & \textcolor{myred}{PV} & 0.5761 & 0.2157 & 0.4510 & 0.2602 & 0.5641 & 0.2472 & 0.6550 & 0.1434 & 0.5160 & 0.2423 \\
 & \cellcolor{yellow!30}\textbf{PSII} & \cellcolor{yellow!30}\textbf{0.2550} & \cellcolor{yellow!30}\textbf{0.0848} & \cellcolor{yellow!30}\textbf{0.2367} & \cellcolor{yellow!30}\textbf{0.1533} & \cellcolor{yellow!30}\textbf{0.3015} & \cellcolor{yellow!30}\textbf{0.1349} & \cellcolor{yellow!30}\textbf{0.2701} & \cellcolor{yellow!30}\textbf{0.0604} & \cellcolor{yellow!30}\textbf{0.2593} & \cellcolor{yellow!30}\textbf{0.1303} \\
\midrule
\textbf{Mistral-24B} & Direct & 0.5491 & 0.1842 & 0.5250 & 0.2922 & 0.6010 & 0.2458 & 0.6226 & 0.1285 & 0.5547 & 0.2504 \\
 & High-Temp & 0.4298 & 0.1381 & 0.3157 & 0.1831 & 0.4625 & 0.1804 & 0.5624 & 0.1109 & 0.3894 & 0.1700 \\
 & Multilingual & 0.4598 & 0.1588 & 0.4035 & 0.2220 & 0.4558 & 0.1776 & 0.5219 & 0.1028 & 0.4343 & 0.1918 \\
 & DivReq & 0.5081 & 0.1615 & 0.4475 & 0.2452 & 0.5455 & 0.2193 & 0.5986 & 0.1113 & 0.4929 & 0.2152 \\
 & PE & 0.4533 & 0.1502 & 0.4722 & 0.2638 & 0.4970 & 0.1936 & 0.6083 & 0.1177 & 0.4813 & 0.2153 \\
 & SimVBG & 0.3184 & \textbf{0.0956} & 0.3052 & 0.1762 & 0.3953 & 0.1592 & 0.4443 & 0.1024 & 0.3386 & 0.1520 \\
 & \textcolor{myred}{PV} & 0.5009 & 0.1851 & 0.4757 & 0.2641 & 0.5496 & 0.2249 & 0.5447 & 0.1169 & 0.5037 & 0.2309 \\
 & \cellcolor{yellow!30}\textbf{PSII} & \cellcolor{yellow!30}\textbf{0.3143} & \cellcolor{yellow!30}0.0984 & \cellcolor{yellow!30}\textbf{0.2577} & \cellcolor{yellow!30}\textbf{0.1641} & \cellcolor{yellow!30}\textbf{0.3553} & \cellcolor{yellow!30}\textbf{0.1479} & \cellcolor{yellow!30}\textbf{0.2990} & \cellcolor{yellow!30}\textbf{0.0645} & \cellcolor{yellow!30}\textbf{0.2971} & \cellcolor{yellow!30}\textbf{0.1419} \\
\bottomrule
\end{tabular}
\end{table*}

\subsection{\textcolor{myred}{Quantitative Comparison Using JS Divergence and MAE}}\label{sec:addres1}

We report both the \textbf{Jensen-Shannon (JS) divergence} and \textbf{Mean Absolute Error (MAE)} for all baseline methods and PSII across multiple models and question categories. Table~\ref{maintab2} summarizes these supplementary results on the WVS dataset. From Table~\ref{maintab2}, we observe that PSII consistently outperforms all baseline methods across all four question categories. In particular, PSII achieves substantial reductions in both JS divergence and MAE, indicating that it generates synthetic populations with distributions that more closely match the real WVS data while maintaining low per-item error.

We report additional ablation study results for PSII using \textbf{JS divergence} and \textbf{MAE} to evaluate the impact of each key component. Table~\ref{ablation2} shows the performance when individual modules are removed. It can be seen that for both metrics, removing any core component leads to performance degradation, further confirming that each module is critical for the overall effectiveness of PSII and the fidelity of the generated distributions.

\begin{table*}
\caption{Ablation study results on PSII across different models. Each row shows the impact of removing one component on JS divergence and MAE. Removing any module results in performance degradation.}
\label{ablation2}
\begin{tabular}{lcccccccc}
\toprule
\multirow{2}{*}{Setting} & \multicolumn{2}{c}{Qwen2.5-7B} & \multicolumn{2}{c}{Qwen2.5-14B} & \multicolumn{2}{c}{Llama-3.1-8B} & \multicolumn{2}{c}{Mistral-24B} \\
\cmidrule(lr){2-3} \cmidrule(lr){4-5} \cmidrule(lr){6-7} \cmidrule(lr){8-9}
 & JS $\downarrow$ & MAE $\downarrow$ & JS $\downarrow$ & MAE $\downarrow$ & JS $\downarrow$ & MAE $\downarrow$ & JS $\downarrow$ & MAE $\downarrow$ \\
\midrule
\cellcolor{yellow!30}PSII (full model) & \cellcolor{yellow!30}\textbf{0.2650} & \cellcolor{yellow!30}\textbf{0.1277} & \cellcolor{yellow!30}\textbf{0.3491} & \cellcolor{yellow!30}\textbf{0.1573} & \cellcolor{yellow!30}\textbf{0.2593} & \cellcolor{yellow!30}\textbf{0.1303} & \cellcolor{yellow!30}0.2971 & \cellcolor{yellow!30}0.1419 \\
PSII w/o value vector & 0.2975 & 0.1373 & 0.3998 & 0.1768 & 0.2807 & 0.1424 & 0.3451 & 0.1603 \\
PSII w/o demographic vectors & 0.4526 & 0.2033 & 0.4368 & 0.1954 & 0.3946 & 0.1739 & 0.4157 & 0.1862 \\
PSII w/o prompt-based profile & 0.2795 & 0.1363 & 0.3678 & 0.1664 & 0.3102 & 0.1617 & 0.3069 & 0.1494 \\
PSII w/o parametric noise & 0.3753 & 0.1778 & 0.4171 & 0.1846 & 0.3478 & 0.1599 & 0.3904 & 0.1753 \\
PSII w/o layer-wise injection & 0.3174 & 0.1523 & 0.3933 & 0.1850 & 0.3137 & 0.1500 & \textbf{0.2922} & \textbf{0.1382} \\
\bottomrule
\end{tabular}
\end{table*}

\subsection{\textcolor{myred}{Layer-Wise Analysis of Demographic Attribute Injection}}\label{sec:addres2}

\begin{figure*}[ht]
  \centering
  \includegraphics[width=\linewidth]{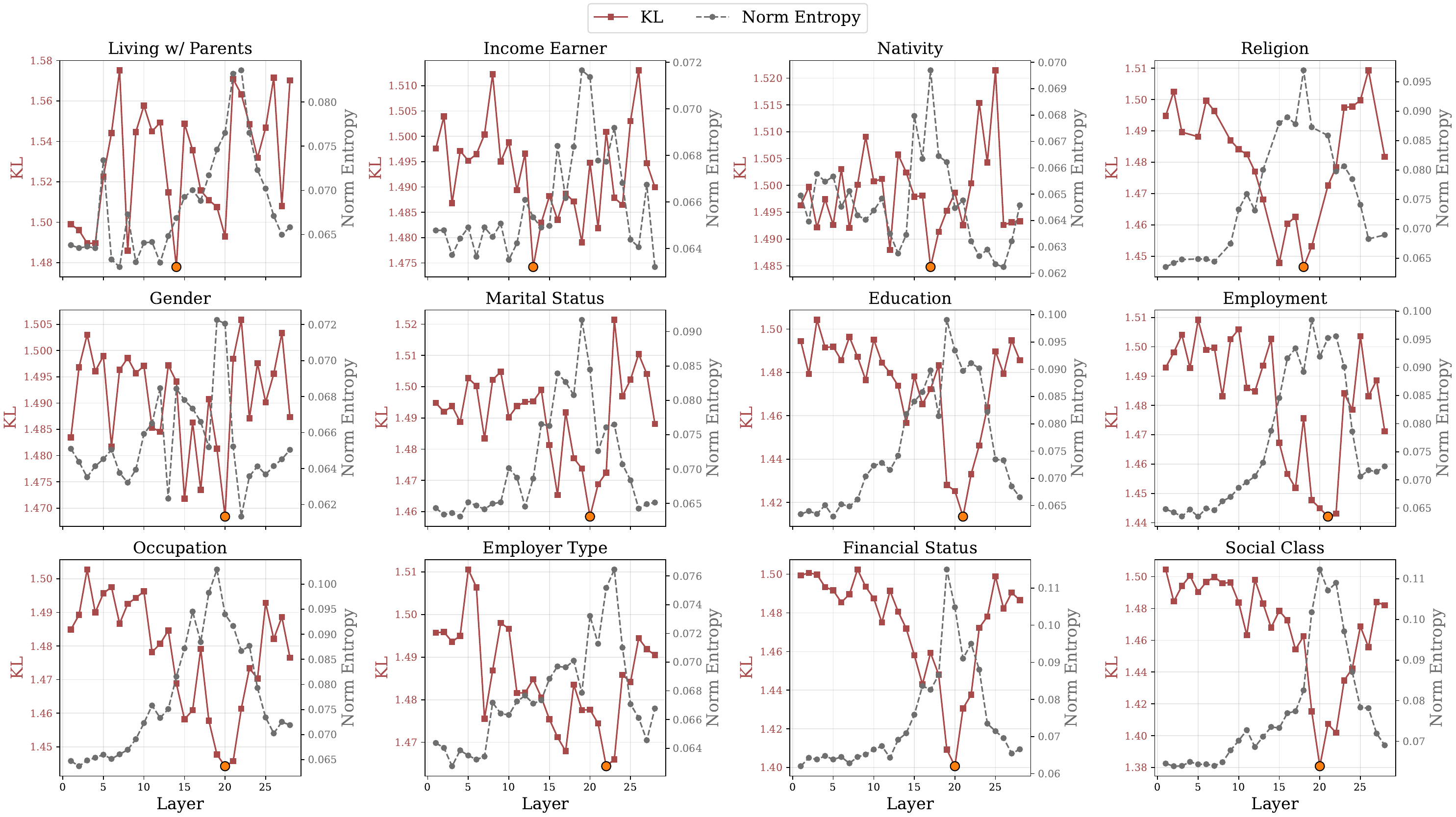}
  \caption{The effects of injecting demographic attributes into different network layers. It illustrates how layer selection impacts simulation accuracy (KL divergence) and diversity (normalized entropy).}
  \Description{Visualization showing which demographic features are injected at low, intermediate, and upper layers, and their impact on KL divergence and normalized entropy.}
  \label{fig:overall_layer}
\end{figure*}

Figure~\ref{fig:overall_layer} illustrates the effects of injecting demographic attributes into different layers of the Transformer network. Each subplot corresponds to a demographic feature, showing how its injection at a specific layer impacts both simulation accuracy, measured by KL divergence, and diversity, measured by normalized entropy. The figure demonstrates that different attributes achieve optimal performance at distinct layers, motivating our hierarchical, layer-wise injection strategy in PSII. By selecting injection points aligned with each attribute’s functional role, we can maximize both accuracy and diversity in the simulated responses.

\subsection{\textcolor{myred}{Layer-Wise Injection Sensitivity Analysis}}\label{sec:addres2.5}

\textcolor{myred}{To further demonstrate that the layer-wise injection strategy in PSII is meaningful rather than heuristic, we conduct additional sensitivity analyses on Qwen2.5-7B beyond the original ablation study. Specifically, we compare the following configurations:}

\begin{itemize}
\item \textbf{\textcolor{myred}{Optimal Layer Configuration (OLC)}}: \textcolor{myred}{The default layer-wise injection strategy used in PSII.}
\item \textbf{\textcolor{myred}{Random Layer Selection (1 \& 2)}}: \textcolor{myred}{Two different random assignments of demographic attributes to layers.}
\item \textbf{\textcolor{myred}{Global Layer Shift (+2 / –2)}}: \textcolor{myred}{Shifting the optimal layer configuration upward or downward by two layers.}
\item \textbf{\textcolor{myred}{Single-Layer Injection (50\% / 70\% / 80\% depth)}}: \textcolor{myred}{Injecting all demographic vectors at a single fixed layer at the specified model depth.}
\end{itemize}

\textcolor{myred}{Table~\ref{tab:layer_sensitivity} reports the KL divergence (lower is better) and Euclidean distance (higher indicates better diversity) for each configuration.}

\begin{table}[H]
\centering
\caption{Layer-wise injection sensitivity analysis on Qwen2.5-7B.}
\label{tab:layer_sensitivity}
\begin{tabular}{lcc}
\toprule
\textbf{Layer Setting} & \textbf{KL $\downarrow$} & \textbf{ED $\downarrow$}\\
\midrule
L1: Optimal Layer Configuration (OLC) & \textbf{0.4843} & \textbf{0.0319} \\
L2: Random Layer Selection (1) & 1.0467 & 0.0812 \\
L3: Random Layer Selection (2) & 1.4093 & 0.2621 \\
L4: Global Layer Shift (+2 from OLC) & 0.5496 & 0.1046 \\
L5: Global Layer Shift (–2 from OLC) & 0.4858 & 0.0852 \\
L6: Single-Layer Injection (50\% depth) & 0.9798 & 0.0367 \\
L7: Single-Layer Injection (70\% depth) & 0.5313 & 0.0964 \\
L8: Single-Layer Injection (80\% depth) & 0.6854 & 0.2018 \\
\bottomrule
\end{tabular}
\end{table}

\textcolor{myred}{The results show that OLC achieves the best overall performance. Configurations with nearby shifts (e.g., global shifts of +2 or –2) remain competitive, while random layer assignments lead to substantial degradation in both accuracy and diversity. Single-layer injection strategies also underperform OLC, with only the 70\% depth configuration approaching but not surpassing OLC's KL performance, while exhibiting worse diversity.}

\textcolor{myred}{These findings confirm that the layer-wise injection strategy is not heuristic but rather a carefully calibrated design that meaningfully contributes to PSII's effectiveness. The optimal assignment of demographic attributes to specific layers matters, and deviations from this configuration result in measurable performance loss.}

\subsection{\textcolor{myred}{Representation-Level Diversity Visualization}}\label{sec:addres3}

\begin{figure}[H]
  \centering
  \includegraphics[width=\linewidth]{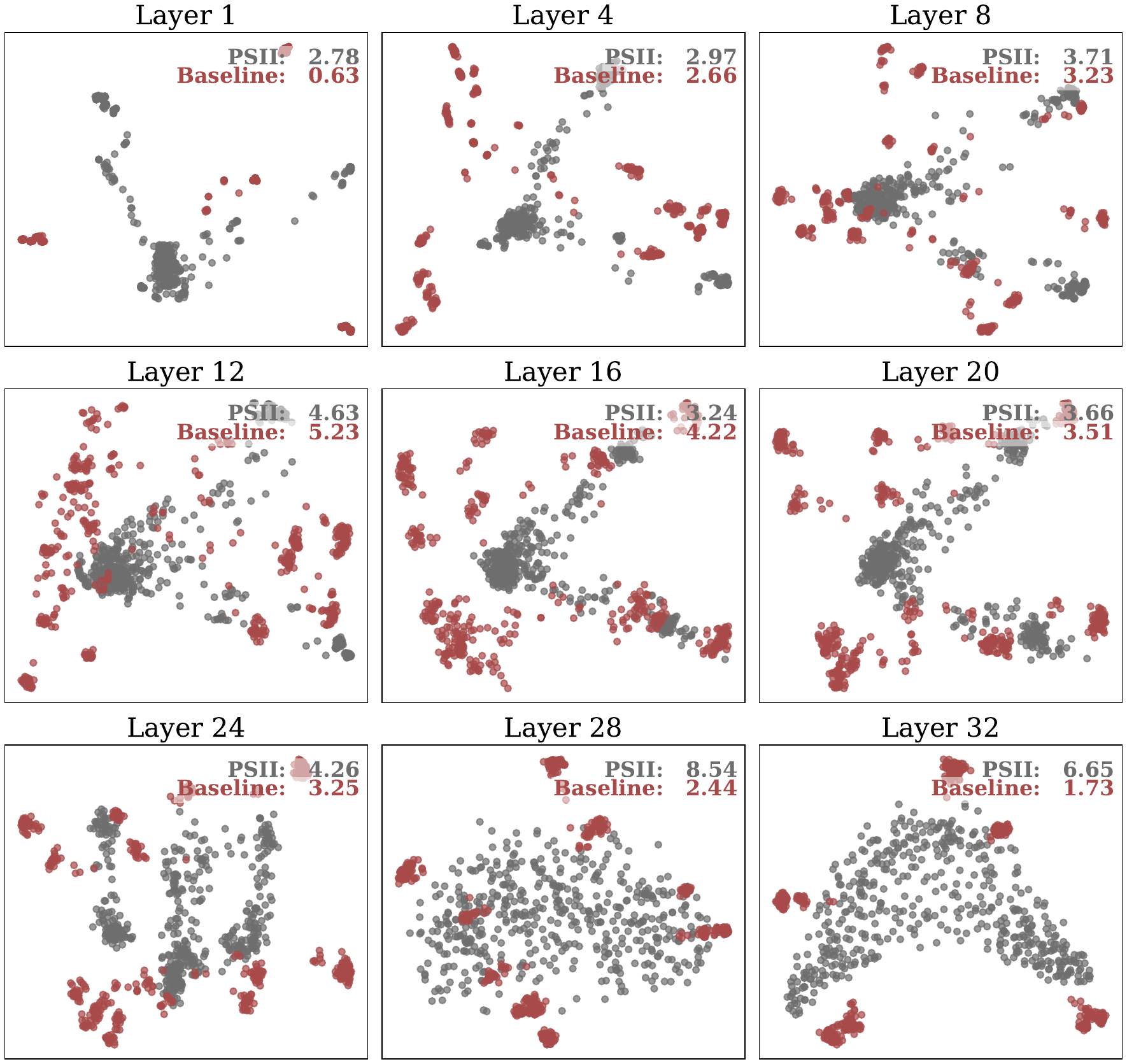}
  \caption{Layer-wise scatter plots of final-token hidden states for 500 simulated agents in \textbf{Llama-3.1-8B}. Red points correspond to baseline methods, while gray points correspond to agents generated using PSII. The reported scores measure the average spatial dispersion of representations in each layer.}
  \label{fig:scatter-llama}
  \Description{Layer-wise scatter plot of final-token hidden states for 500 simulated agents. Red points correspond to baseline methods, while gray points correspond to agents generated using PSII.}
\end{figure}

\begin{figure}[H]
  \centering
  \includegraphics[width=\linewidth]{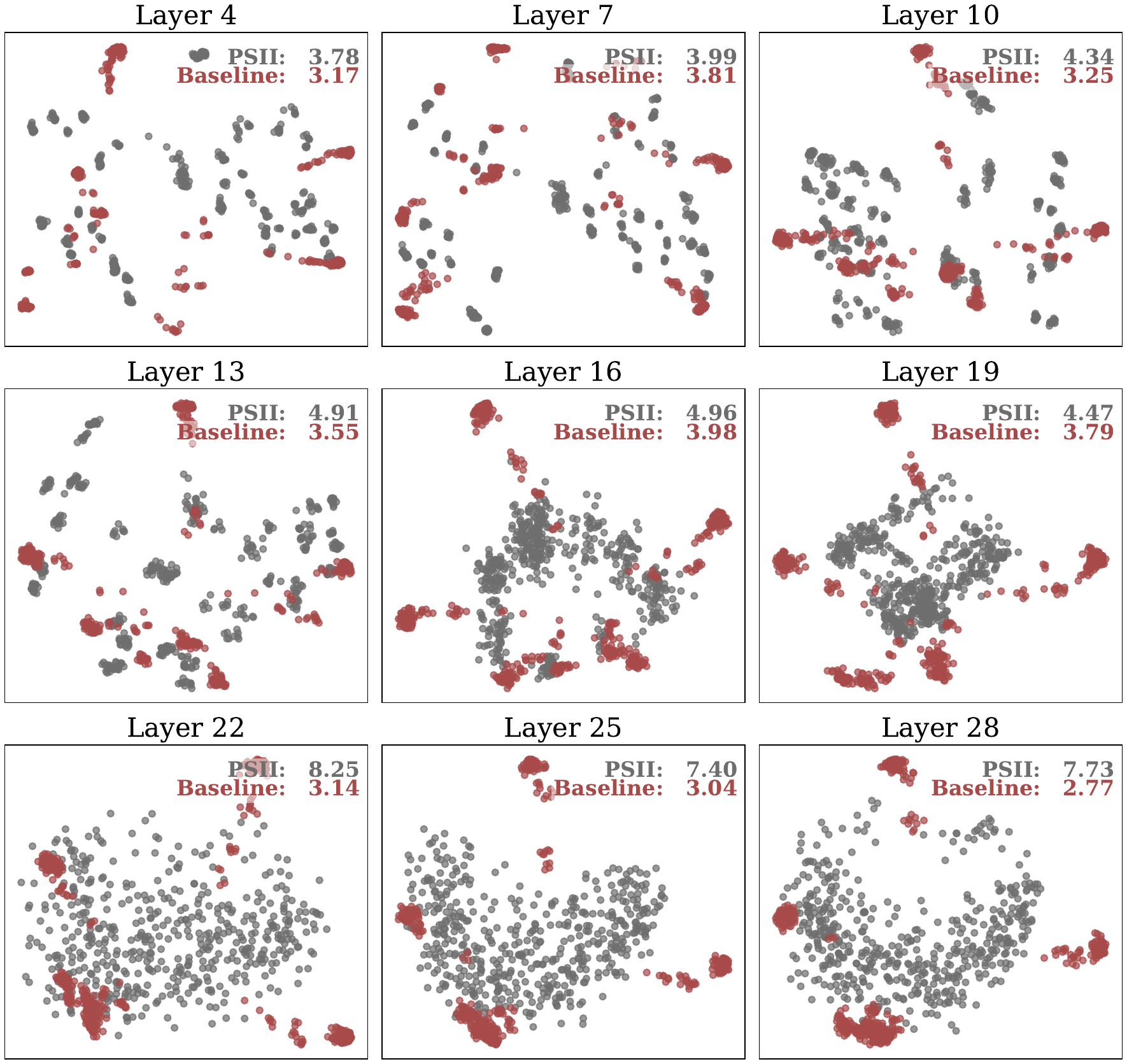}
  \caption{Layer-wise scatter plots of final-token hidden states for 500 simulated agents in \textbf{Qwen2.5-7B}. Red points correspond to baseline methods, while gray points correspond to agents generated using PSII. The reported scores measure the average spatial dispersion of representations in each layer.}
  \label{fig:scatter-qwen7}
  \Description{Layer-wise scatter plot of final-token hidden states for 500 simulated agents. Red points correspond to baseline methods, while gray points correspond to agents generated using PSII.}
\end{figure}

\begin{figure}[H]
  \centering
  \includegraphics[width=\linewidth]{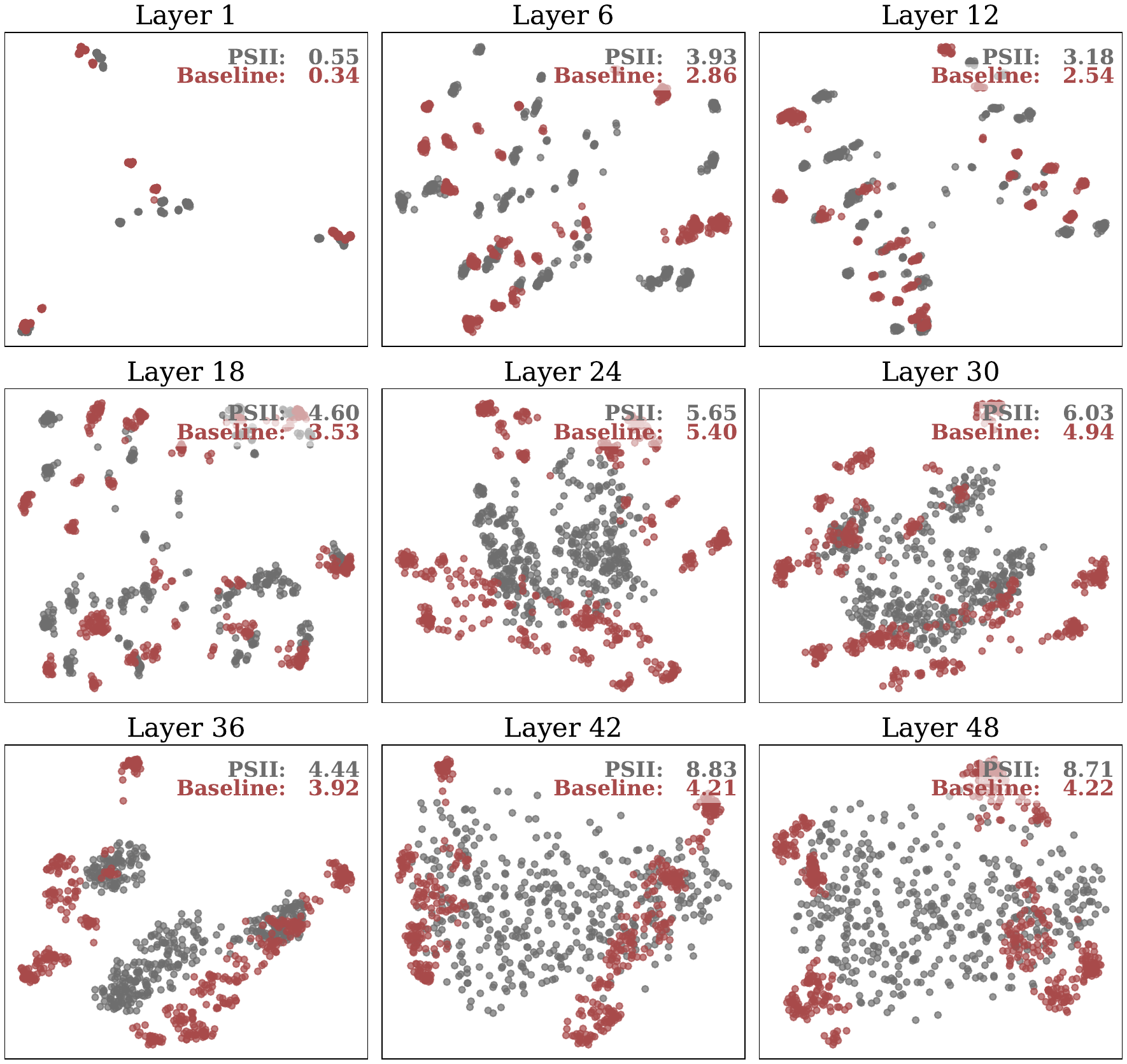}
  \caption{Layer-wise scatter plots of final-token hidden states for 500 simulated agents in \textbf{Qwen2.5-14B}. Red points correspond to baseline methods, while gray points correspond to agents generated using PSII. The reported scores measure the average spatial dispersion of representations in each layer.}
  \label{fig:scatter-qwen14}
  \Description{Layer-wise scatter plot of final-token hidden states for 500 simulated agents. Red points correspond to baseline methods, while gray points correspond to agents generated using PSII.}
\end{figure}

\begin{figure}[H]
  \centering
  \includegraphics[width=\linewidth]{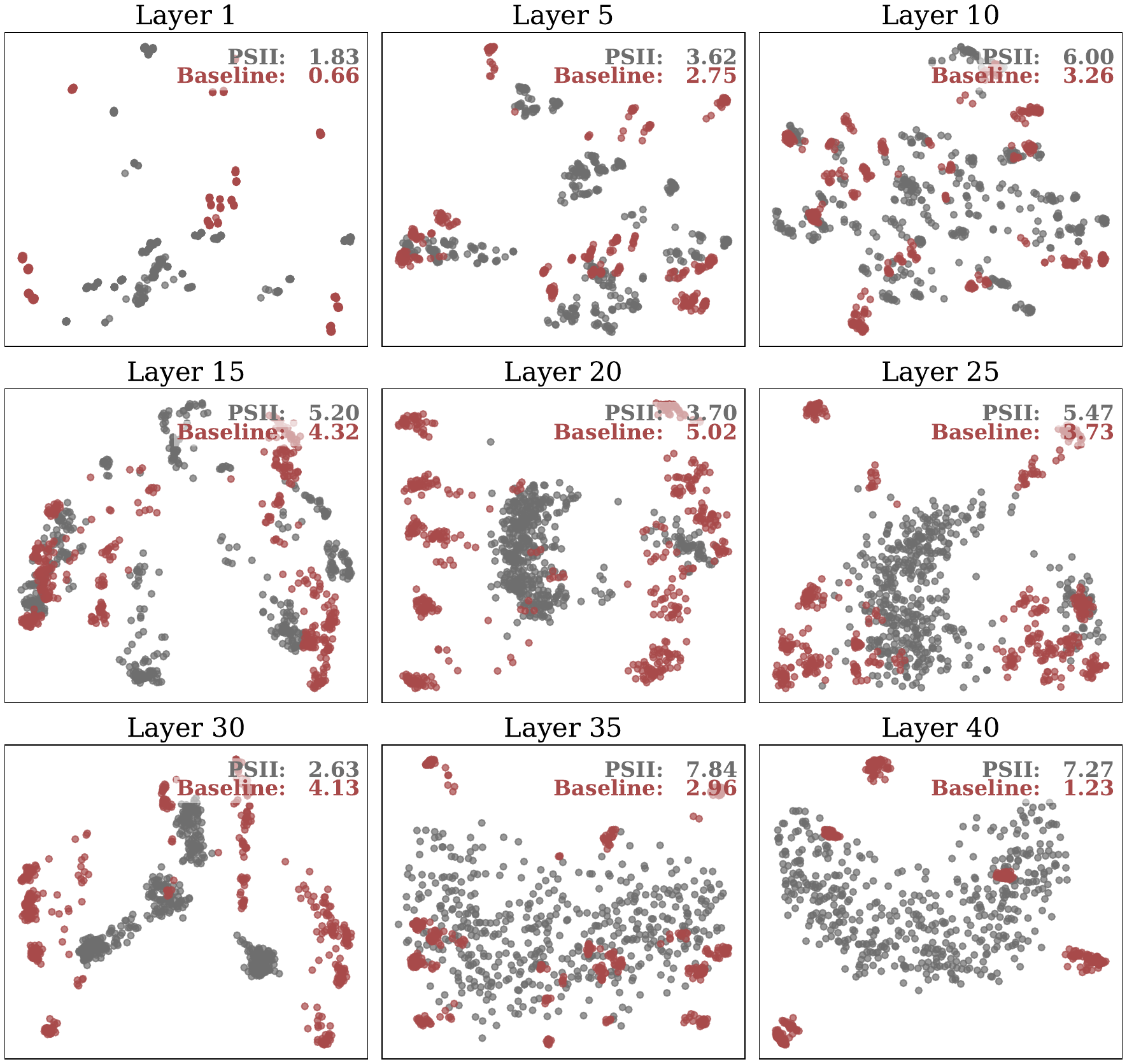}
  \caption{Layer-wise scatter plots of final-token hidden states for 500 simulated agents in \textbf{Mistral-24B}. Red points correspond to baseline methods, while gray points correspond to agents generated using PSII. The reported scores measure the average spatial dispersion of representations in each layer.}
  \label{fig:scatter-mistral}
  \Description{Layer-wise scatter plot of final-token hidden states for 500 simulated agents. Red points correspond to baseline methods, while gray points correspond to agents generated using PSII.}
\end{figure}

To further illustrate how PSII improves population heterogeneity at the representation level, Figures~\ref{fig:scatter-llama}--\ref{fig:scatter-mistral} show layer-wise scatter plots of the final-token hidden states for 500 simulated agents across four different LLMs: Qwen2.5-7B, Qwen2.5-14B, Llama-3.1-8B, and Mistral-24B. \textcolor{myred}{We compare the baseline (prompt engineering + multilingual) with PSII, and visualize hidden states for a randomly selected question (Q112) via KPCA; each point represents an agent.} Red points correspond to baseline methods, while gray points correspond to agents generated using PSII. We further quantify representation-level heterogeneity using a k-nearest-neighbor (kNN) radius metric, defined as the average distance from each hidden-state vector $\mathbf{h}_i$ to its $k$-th nearest neighbor, scaled by a factor of 100 for readability, where larger values indicate more dispersed and diverse representations. These visualizations indicate that baseline methods tend to exhibit clustering and lack of diversity in the higher-layer hidden states, manifesting the so-called \textbf{Diversity Collapse} phenomenon. In contrast, PSII maintains a more dispersed and structured distribution, better capturing the underlying heterogeneity in demographic and value attributes. Notably, this pattern is consistent across all four models, demonstrating the robustness of the PSII approach.

Overall, these additional results reinforce the main findings: PSII not only improves distributional fidelity and per-item accuracy compared to baseline methods, but also preserves diversity and heterogeneity in the model’s internal representations, which is critical for realistic population simulation.

\end{document}